%% file: ijcai24.tex
\pdfoutput=1

\documentclass{article}
\usepackage{ijcai24}

\usepackage{times}
\usepackage{soul}
\usepackage{url}
\usepackage[hidelinks]{hyperref}
\usepackage[utf8]{inputenc}
\usepackage[small]{caption}
\usepackage{graphicx}
\usepackage{amsmath}
\usepackage{amsthm}
\usepackage{booktabs}
\usepackage{algorithm}
\usepackage{algorithmic}
\usepackage[switch]{lineno}

\usepackage{CJKutf8}
\newcommand{\cnw}[1]{\begin{CJK}{UTF8}{gbsn}#1\end{CJK}}
\usepackage{color}
\usepackage{MnSymbol}
\usepackage{tabularx}
\usepackage{pifont}
\usepackage{newunicodechar}
\newunicodechar{✓}{\ding{51}}
\newunicodechar{✗}{\ding{55}}
\usepackage{booktabs}
\usepackage{multirow}
\newif\ifdraft
\drafttrue
\draftfalse

\newcommand{\comment}[1]{}
\newcommand{\jy}[1]{\ifdraft {\textcolor{blue}{#1}} \else {#1}\fi} 
\newcommand{\sg}[1]{\ifdraft {\textcolor{magenta}{#1}} \else {#1}\fi} 

\newcommand{\ie}{\textit{i}.\textit{e}., }
\newcommand{\eg}{\textit{e}.\textit{g}.\ }

\usepackage{subcaption}
\usepackage{enumitem}
\setlist[enumerate]{leftmargin=0pt}
\usepackage{mathtools}

\title{Efficient and Scalable Chinese Vector Font Generation via Component Composition}

\author{
Jinyu Song$^1$\and
Weitao You$^1$\and
Shuhui Shi$^1$\and
Shuxuan Guo\and
Lingyun Sun$^1$\And
Wei Wang$^2$\footnote{Corresponding author.}\\
\affiliations
$^1$ College of Computer Science and Technology, Zhejiang University \\
$^2$ Beijing Jiaotong University \\
\emails
\{songjinyu,weitao\_you,shuhui.shi,sunly\}@zju.edu.cn,
gsxuan.work@gmail.com,
wei.wang@bjtu.edu.cn
}

\begin{document}

\maketitle

\begin{abstract}
Chinese vector font generation is challenging due to the complex structure and huge amount of Chinese characters. Recent advances remain limited to generating a small set of characters with simple structure. In this work, we first observe that most Chinese characters can be disassembled into frequently-reused components. Therefore, we introduce the first efficient and scalable Chinese vector font generation approach via component composition, allowing generating numerous vector characters from a small set of components. To achieve this, we collect a large-scale dataset that contains over \textit{90K} Chinese characters with their components and layout information. Upon the dataset, we propose a simple yet effective framework based on spatial transformer networks (STN) and multiple losses tailored to font characteristics to learn the affine transformation of the components, which can be directly applied to the Bézier curves, resulting in Chinese characters in vector format. Our qualitative and quantitative experiments have demonstrated that our method significantly surpasses the state-of-the-art vector font generation methods in generating large-scale complex Chinese characters in both font generation and zero-shot font extension. 
\end{abstract}

\input{intro}

\input{related-work}
\input{dataset}
\input{method}

\input{evaluation}
\input{conclusion}
\bibliographystyle{named}
\bibliography{ijcai24}

\appendix
\input{supp-comparison}
\input{supp-dataset}
\input{supp-method}

\input{supp-eval}

\input{supp-impact}

\end{document}

%% file: intro.tex

\section{Introduction}

\begin{figure}
\centering
\begin{subfigure}[b]{0.73\linewidth}
         \centering
         \includegraphics[width=\columnwidth]{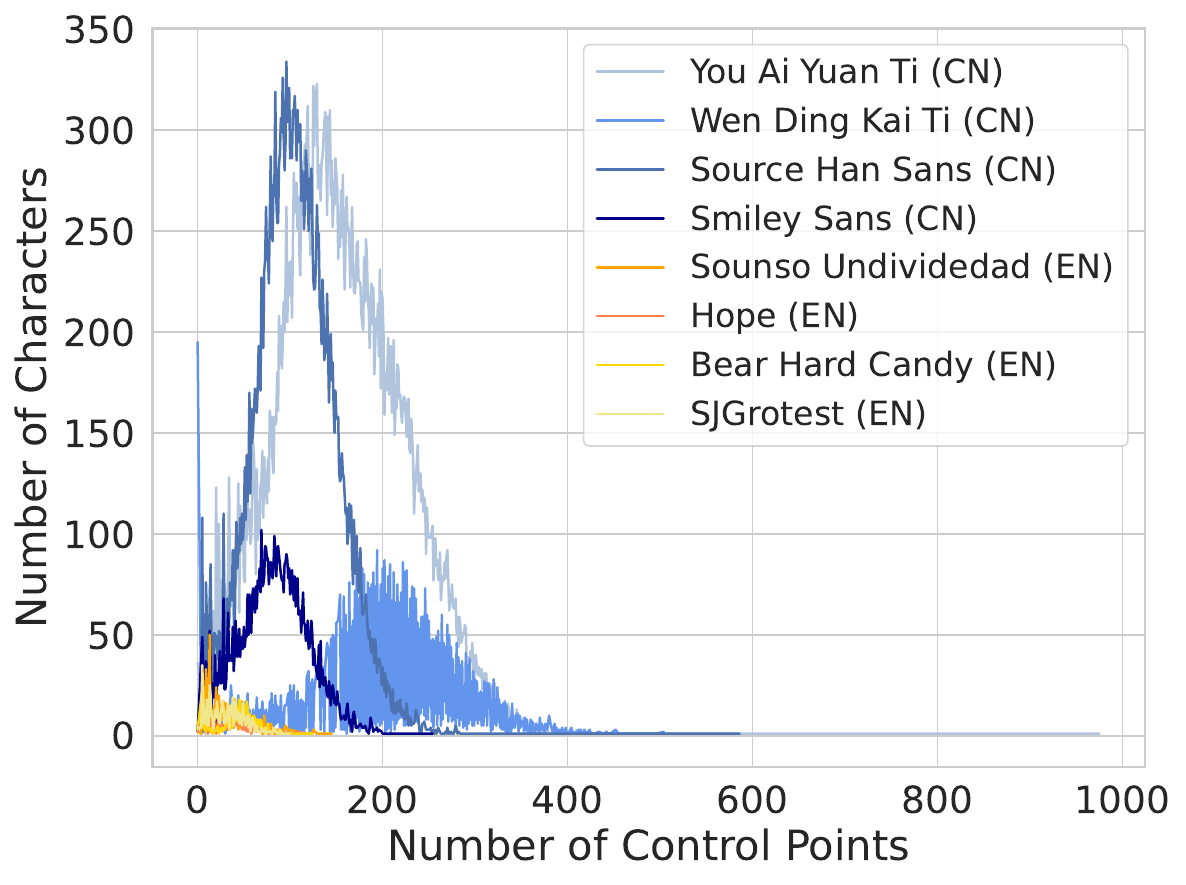}
 \end{subfigure}
\begin{subfigure}[b]{0.25\linewidth}
    \centering
    \includegraphics[width=\columnwidth]{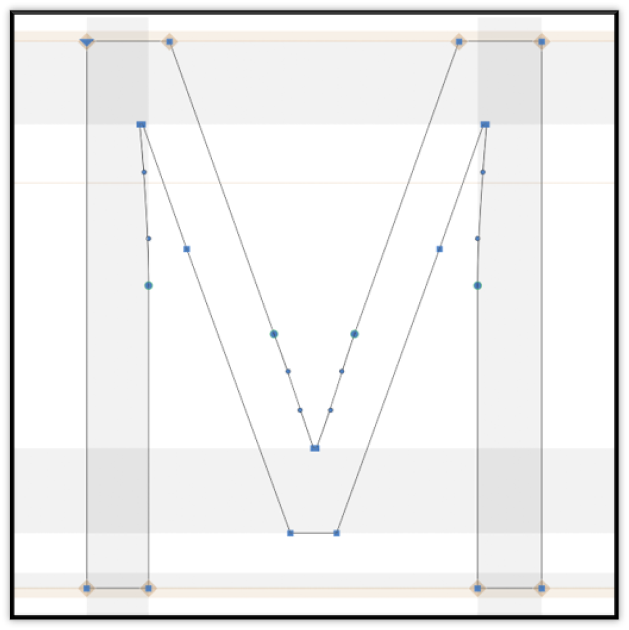} \\
    \vspace{0.25cm}
    \includegraphics[width=\columnwidth]{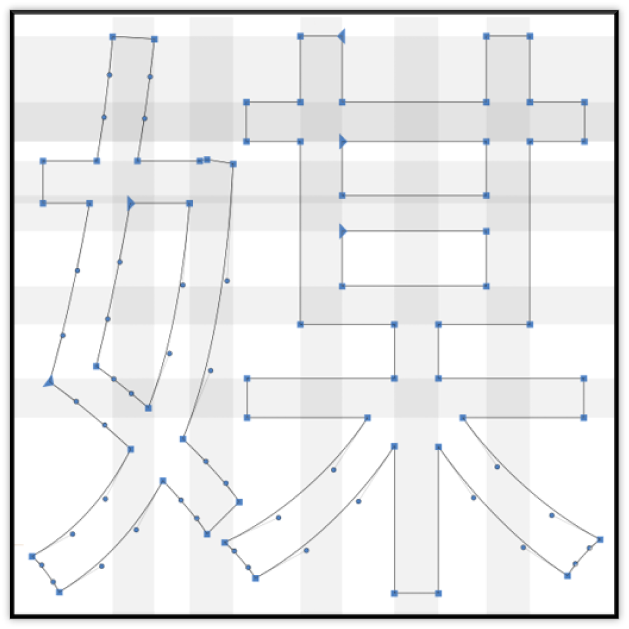}
			
\end{subfigure}
\caption{
\textbf{Comparison of control points between Chinese and English fonts.} \normalfont{\textbf{Left}:The dark lines represent Chinese fonts, while the warm lines are English fonts. \textbf{Right}: Examples of control points for English character "M" (Top) and Chinese character "\cnw{媒}" (Bottom). 
The number of control points per character in the Chinese fonts is much \sg{larger} than that in the English fonts, indicating that Chinese fonts are more complex than English ones. }
}
\vspace{-1.5mm}
\label{fig:dist_control_points_characters}
\end{figure}

Chinese font design has been acknowledged as a notoriously challenging task due to the complexity and huge amount of Chinese characters. 
The vast number of characters (\textit{90K}, according to GB18030-2022~\footnote{The GB18030-2022 standard is the current largest character set of Chinese characters.} standard) makes it time-consuming for designers to design fonts covering all characters.
Moreover, as hieroglyphics, Chinese characters tend to be more intricate than characters in other languages, such as English, requiring designers to use more control points (\ie Bézier curves) to outline them precisely, as shown in Fig.~\ref{fig:dist_control_points_characters}.\comment{, Chinese fonts in general require a greater number of control points per character compared to English fonts.}
\comment{As a consequence, most Chinese fonts in the market only covers 6,763 characters to save costs and ensure the basic availability. However, this may lead to the incorrect display of unsupported characters, resulting in inconsistent content or style, which in turn diminishes the user experience.} 
These challenges have prompted the need for efficient and scalable approaches to Chinese vector font generation to accelerate Chinese font design process.


Recent advances in font generation have made significant progress in both image-based~\cite{lyu2017auto,tian2017zi2zi,zhang2018separating,sun2018pyramid,jiang2019scfont,zhang2020ssnet,wang2020attribute2font,park2021multiple,kong2022look} and vector-based~\cite{lopes2019learned,carlier2020deepsvg,wang2022aesthetic,wang2023deepvecfont} font generation. Despite achieving improvements in quality and efficiency, the image-based methods remain limitations to accelerate font design due to their \sg{output image} format. In contrast, vector-based approaches \sg{appears to} truly boost design process. However, most research on vector-based font generation focuses on vector graphics and alphabet fonts. Although there have been successful attempts at generating Chinese vector fonts~\cite{wang2023deepvecfont,aoki2022svg}, these methods can only produce a small subset of Chinese characters with simple structure and struggle to generate complex (\ie multiple-component) ones in high quality. Since complex characters constitute the vast majority of all characters (see Fig.\ref{fig:dist_layout}), current vector-based methods are not effective enough to achieve large-scale Chinese vector font generation.
To overcome these challenges, we propose a novel framework for efficient and scalable Chinese vector font generation via component composition. Our approach is motivated by the observation that Chinese characters typically consist of various components. These components are frequently reused, allowing the generation of numerous characters from a small set of components.
In our method, we first construct a large-scale Chinese font-components dataset containing over \textit{90K} characters with their corresponding components and layouts. \sg{To the best of our knowledge, it offers the broadest coverage of Chinese characters}. Based on the dataset, we propose a simple yet efficient model, named ~\textbf{C}omponent \textbf{A}ffine transformation \textbf{R}egressor (CAR)
\jy{to model the Chinese vector font generation as a regression task.} In particular, during training, our CAR
learns affine transformation matrices for character components from a small set of well-pre-designed character images. At inference, the CAR can output the corresponding affine transformation matrices for the components of Chinese characters, which could be applied to the Bézier curves of the components themselves to generate the corresponding un-designed characters in vector format. \sg{To this end, our method is capable of generating large number of complex Chinese vector characters.} Additionally, we develop a Glyphs plugin to integrate our method into the font design pipeline. We conduct extensive quantitative and qualitative evaluations to validate the effectiveness of our method.
In summary, our contribution are as follows:
\begin{itemize}[leftmargin=0.4cm]
  \item We build a \textit{large-scale} Chinese font-components dataset, which contains the components and layout information of more than \textit{90K} Chinese characters. To the best of our knowledge, this is the \textit{first} and \textit{largest} Chinese font-components dataset. \sg{It is note that our dataset could serve as a new benchmark of Chinese vector font generation.}
  \item \sg{We propose a simple yet efficient framework utilizing affine transformation for \textit{large-scale}, \textit{vector-format} Chinese font generation. Our method not only surpasses the state-of-the-art techniques in generating complex Chinese characters, but also 
  enables extending existing fonts set in a zero-shot manner.}
  \item \sg{We develop a Glyphs plugin incorporating our method into the professional workflow of Chinese font design, significantly accelerating Chinese font design.}
\end{itemize}

%% file: related-work.tex
\section{Related Work}

Chinese font generation has been a longstanding topic of research in the fields of Computer Vision and Artificial Intelligence. Early studies primarily focused on handcrafted approaches, such as stroke models~\cite{strassmann1986hairy,guo1991modeling,lee1999simulating}, stroke extraction, and composition~\cite{xu2005automatic,zhou2011easy}. ~\cite{wong1995designing} proposed a components-based Chinese font composition approach that leverages handcrafted rules to generate characters. ~\cite{suveeranont2010example} and ~\cite{campbell2014learning} explored to generate new font by blending the exists fonts.

Afterwards, with the rapid development of generative models such as VAEs~\cite{kingma2013auto}, GANs~\cite{goodfellowNIPS2014generative} and diffusion models~\cite{pmlr-v37-sohl-dickstein15,ho2020denoising}, another family of font generation approaches~\cite{lyu2017auto,tian2017zi2zi,guo2018creating,sun2018pyramid,gao2019artistic,jiang2019scfont} have been proposed.~\cite{lyu2017auto,tian2017zi2zi} treated font generation task as an image-to-image translation task and synthesized Chinese calligraphy font with pix2pix~\cite{isola2017image}. Many studies have followed this thread, making progress by integrating more precise information~\cite{zhang2020ssnet,wang2020attribute2font,park2021multiple}, improving generator~\cite{zhang2018separating,sun2018pyramid,gao2019artistic,park2021multiple}, and discriminator~\cite{kong2022look} architecture. Recently, FontTransformer~\cite{liu2023fonttransformer} and Diff-Font~\cite{he2022diff} employed stacked transformers~\cite{vaswani2017attention} and diffusion model, respectively, for few-shot font image generation. However, these methods still produce image format outputs, which may not directly assist font design or be suitable for printing.

To this end, some recent studies have aimed to directly generate vector fonts, which is a more challenging but promising approach to truly accelerate font design. SVG-VAE~\cite{lopes2019learned}, DeepSVG~\cite{carlier2020deepsvg}, and Im2Vec~\cite{reddy2021im2vec} focused on vector graphics and alphabet fonts. As for Chinese vector font generation~\cite{gao2019automatic,lian2022cvfont,wang2022aesthetic,xia2023vecfontsdf,wang2023deepvecfont},~\cite{zhang2017drawing} and \cite{tang2019fontrnn} proposed to generate writing trajectory of Chinese characters with RNN.\jy{~\cite{wang2022aesthetic} regressed layout from glyphs to generate text logo.}~\cite{aoki2022svg} employed transformers with Chamfer distance~\cite{fan2017point} to generate sans-serif style Chinese vector fonts in SVG format~\footnote{\href{https://en.wikipedia.org/w/index.php?title=SVG&oldid=1149117185}{SVG --- Wikipedia}}. \jy{DeepVecFont-v2~\cite{wang2023deepvecfont} (DVF-v2) improved the performance of DeepVecFont~\cite{wang2021deepvecfont} (DVF) by replacing the RNN with transformer and sampling auxiliary points.}However, the aforementioned methods of Chinese vector font generation are limited to a small set of simple characters and struggle to generate complex Chinese characters multiple paths and long drawing command sequences~\cite{wang2023deepvecfont}. \jy{In contrast, our work is able to generate more than 75,000 Chinese characters in vector format with much fewer components,
enabling efficient and large-scale Chinese vector font generation to accelerate Chinese font design. We also provide a detailed comparison with current vector-based methods in our supplementary material.}


%% file: dataset.tex
\section{Our Large-scale Chinese Font-Components Dataset}
\label{sec:dataset}

In the realm of previous research on Chinese font generation~\cite{lyu2017auto,tian2017zi2zi,jiang2019scfont,park2021multiple,liu2023fonttransformer,he2022diff}, 
including both image-based and vector-based approaches, has been limited to generating no more than 6,763 Chinese characters defined in the GB2312~\footnote{\url{https://en.wikipedia.org/wiki/GB_2312}} character set, while the total number of Chinese characters exceeds 90K~\footnote{\url{https://www.unicode.org/charts/}}. To address these challenges, our work takes advantage of the component-based structure of Chinese characters to generate vector-based, designer-editable character sets. To this end, we have constructed a comprehensive dataset of Chinese characters and their components, covering over 90K Unicode characters, including rare ones. This dataset allows our proposed algorithm to generate a more substantial volume of Chinese characters and address the issue of missing characters in ancient texts that negatively impact reading experience.

\subsection{Preliminary: Component and Layout}
\label{sec:comp_and_layout}

Chinese characters are typically composited by different components, which themselves are also Chinese characters. These components are often reused many times. Note that Chinese character radicals are natural subsets of our components. For the difference between the radical and component, please refer to our supplementary material.

\begin{figure}
    \centering
    \includegraphics[width=0.72\linewidth]{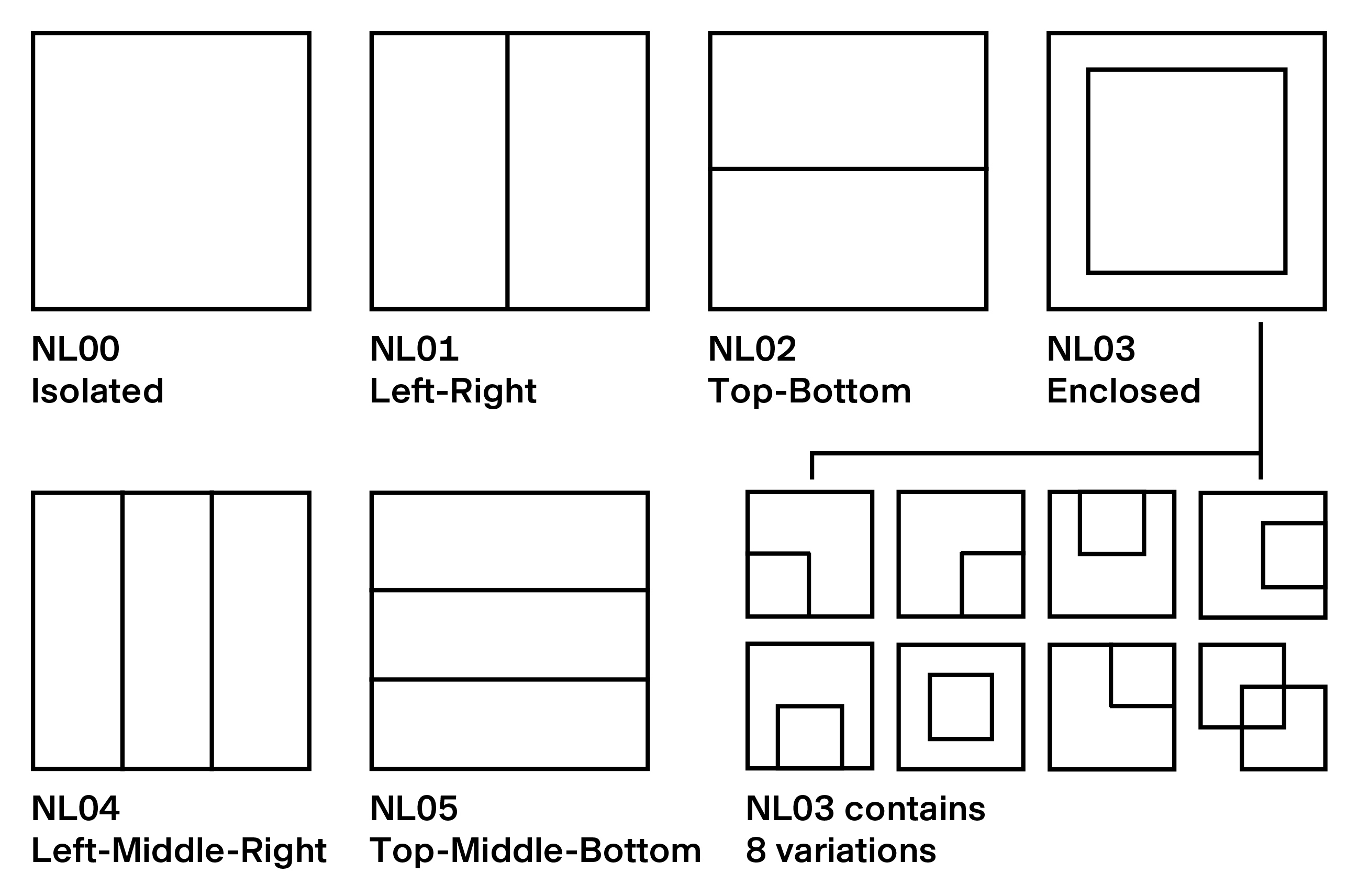}
    \caption{\textbf{6 main layouts for Chinese characters in our dataset.} \normalfont{Note that NL03 has 8 variations and they are categorized as one}.}
    \label{fig:layout}
\end{figure}

Based on these components, \sg{and inspired by the Unicode Ideographic Description Characters~\cite{unicode9.0}}, we then categorized 
Chinese characters into six layouts, as depicted in Fig.~\ref{fig:layout}: NL00 represents Isolated; NL01 denotes Left-Right; NL02 corresponds to Top-Bottom; NL03 signifies Enclosed; NL04 refers to Left-Middle-Right; and NL05 indicates Top-Middle-Bottom. Notably, NL03 encompasses eight distinct variations or sub-layouts. 


\subsection{Our Dataset}
Existing datasets, such as IDS~\footnote{\url{https://github.com/cjkvi/cjkvi-ids}} and other open-source alternatives fails providing the decomposition information for all Unicode-encoded Chinese characters. Therefore, we opted to create a Chinese character component decomposition dataset based on Unicode in order to learn from a diverse range of Chinese character forms.

\begin{figure}
    \centering
    \includegraphics[width=0.95\linewidth]{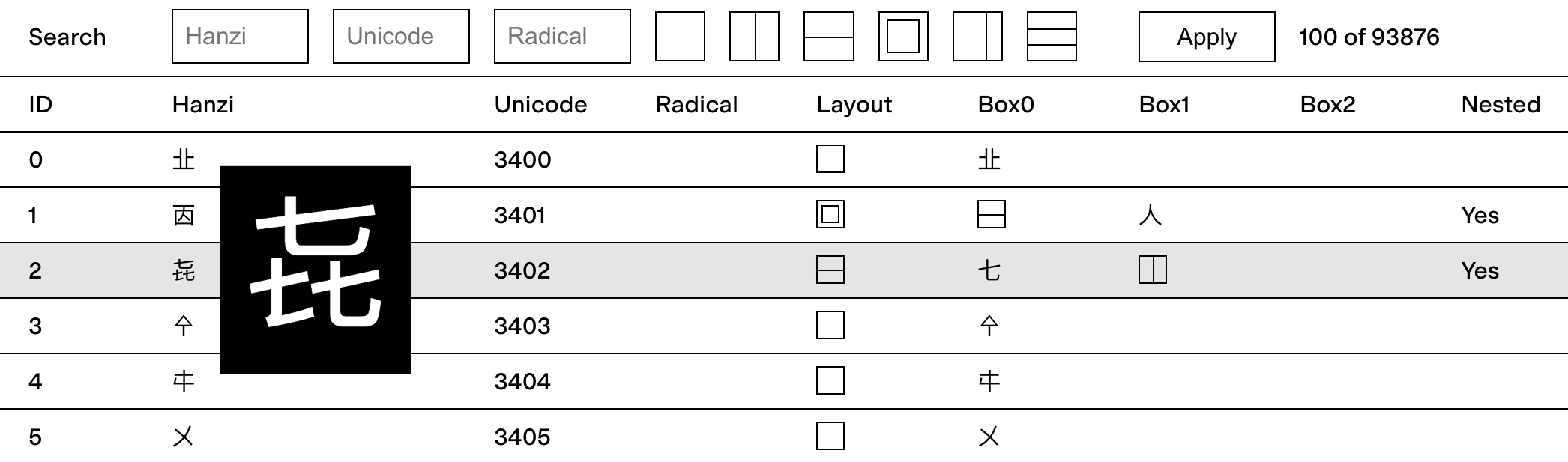}
    \caption{\textbf{Our Dataset.} \normalfont{Each sample contains ID, the character (Hanzi), Unicode, Radicals, Layout and its components.}}
    \label{fig:database}
\end{figure}

The Unicode-encoded Chinese characters are organised sequentially according to their Unicode order, resulting in \textbf{92,560} characters. As depicted in Fig.~\ref{fig:database}, the dataset's first column represents the character ID; the second column shows the character itself; the third column lists the character's Unicode; the fourth column provides component information (if individual radicals are already indicated in some characters' data, they will be listed here for reference); the fifth column describes the character's layout; and the sixth to eighth columns present the decomposed components.

Note that some Chinese characters uses more than two components. When characters require the combination of more components, we nest different structures together and annotate them as "Nested." As shown in Fig.~\ref{fig:database}, the character with the ID 2 has a Left-Right layout nested in the Top-Bottom Layout. For more details of construction of our dataset, please refer to the supplementary material.

\begin{figure*}
     \centering
     \begin{subfigure}[b]{0.231\textwidth}
         \centering
         \includegraphics[width=\linewidth]{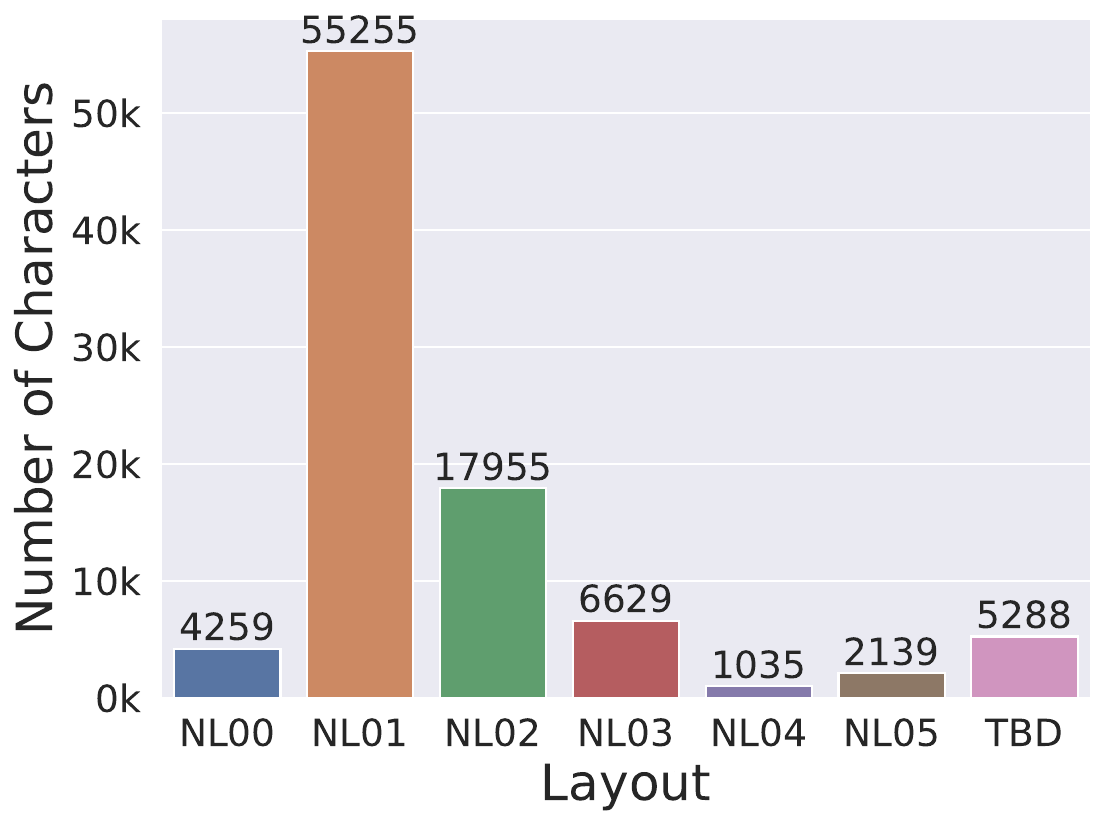}
         \caption{$\#$chars in each layout.}
         \label{fig:dist_layout}
     \end{subfigure}
     \hfill
     \begin{subfigure}[b]{0.231\textwidth}
         \centering
         \includegraphics[width=\linewidth]{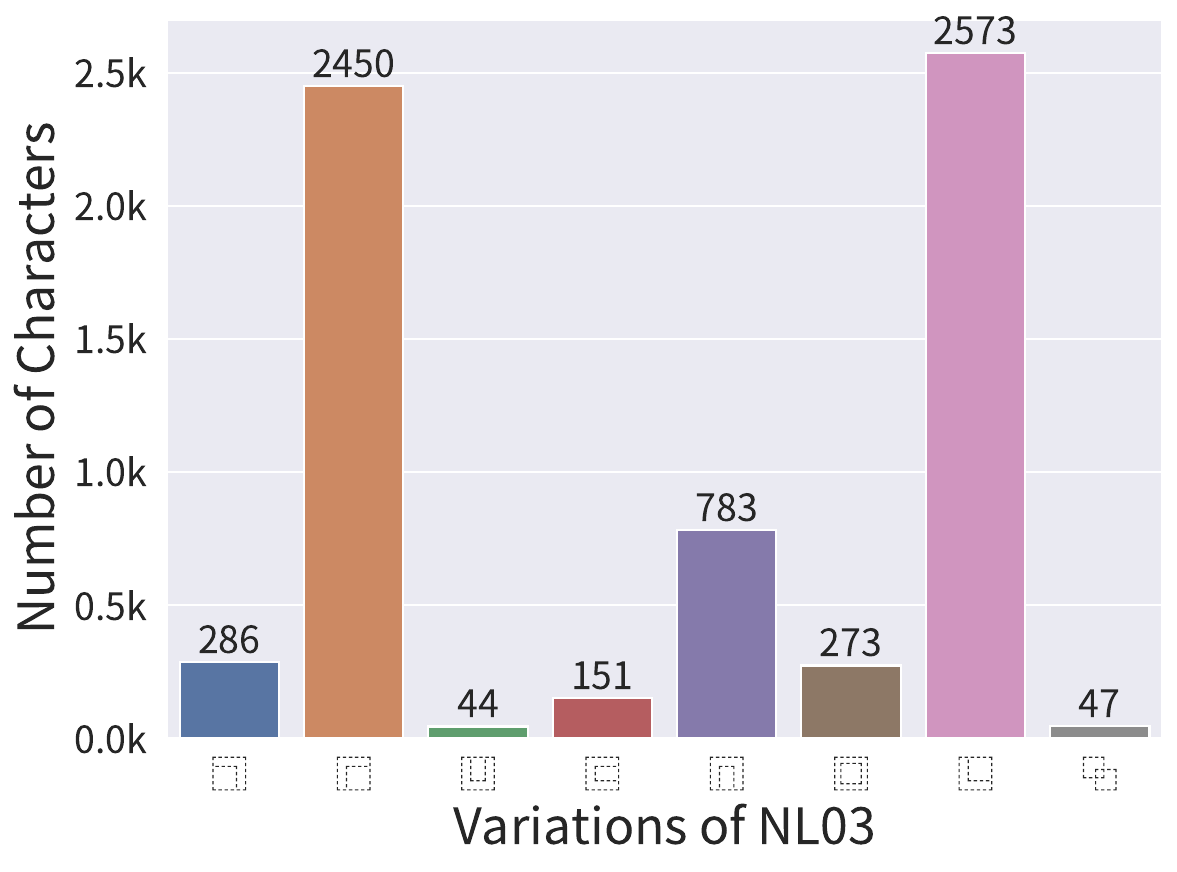}
         \caption{$\#$chars of NL03 variations.}
         \label{fig:sublayout}
     \end{subfigure}
     \hfill
     \begin{subfigure}[b]{0.231\textwidth}
         \centering
         \includegraphics[width=\linewidth]{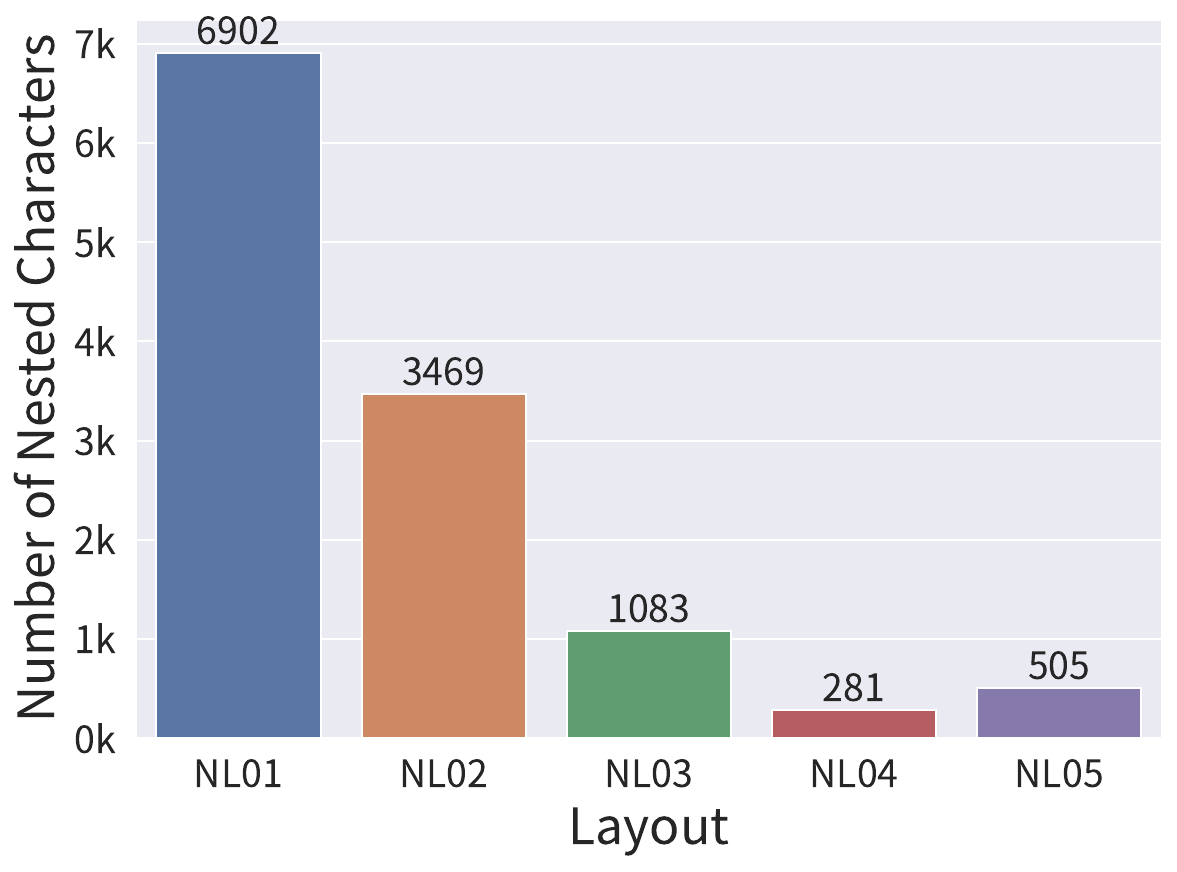}
         \caption{Distribution of nested chars.}
         \label{fig:nested_dist}
     \end{subfigure}
     \begin{subfigure}[b]{0.231\textwidth}
        \centering
        \includegraphics[width=\linewidth]{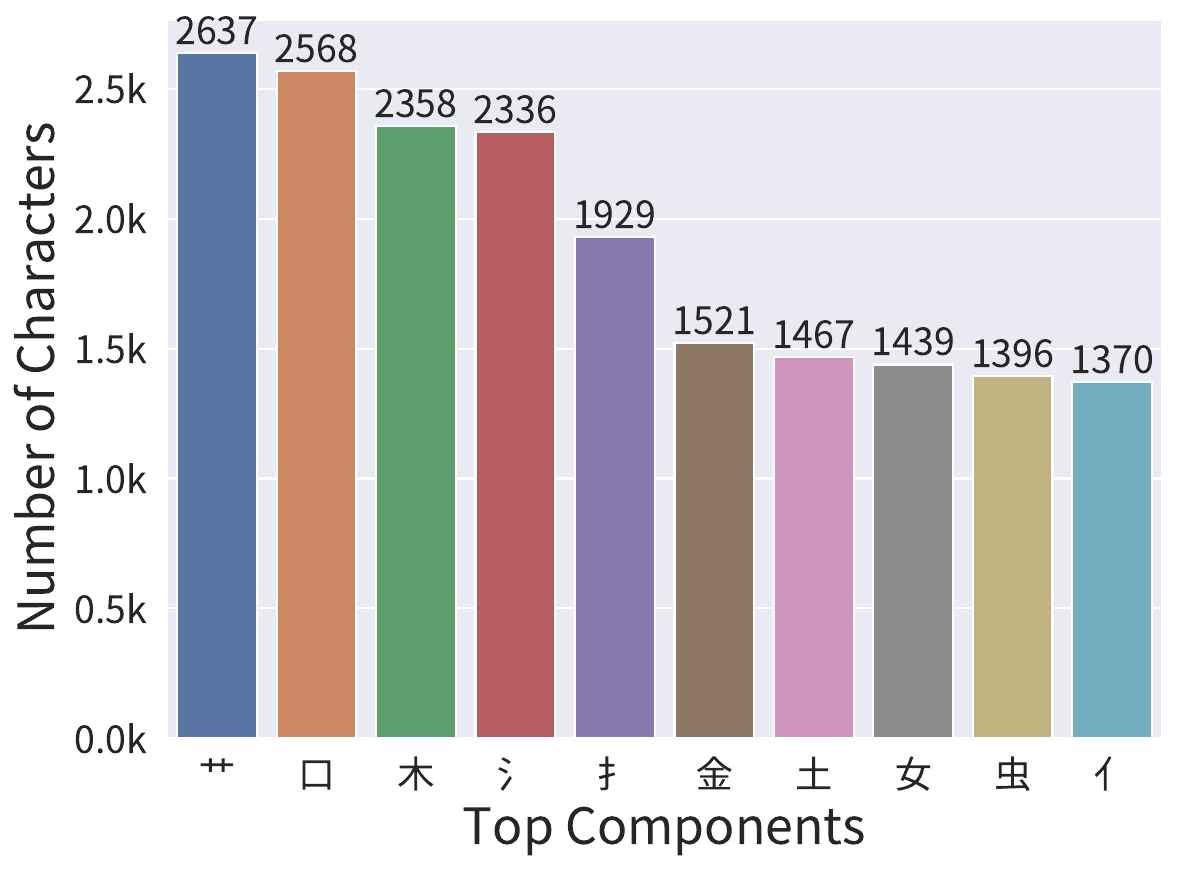}
        \caption{Top 10 Components. 
        }
        \label{fig:top_components}
     \end{subfigure}
        \caption{Statistical Information of our dataset.}
        \label{fig:info_dataset}
    \vspace{-1mm}
\end{figure*}

\subsection{Statistics of the dataset}

\begin{figure}
    \centering
    \includegraphics[width=0.75\linewidth]{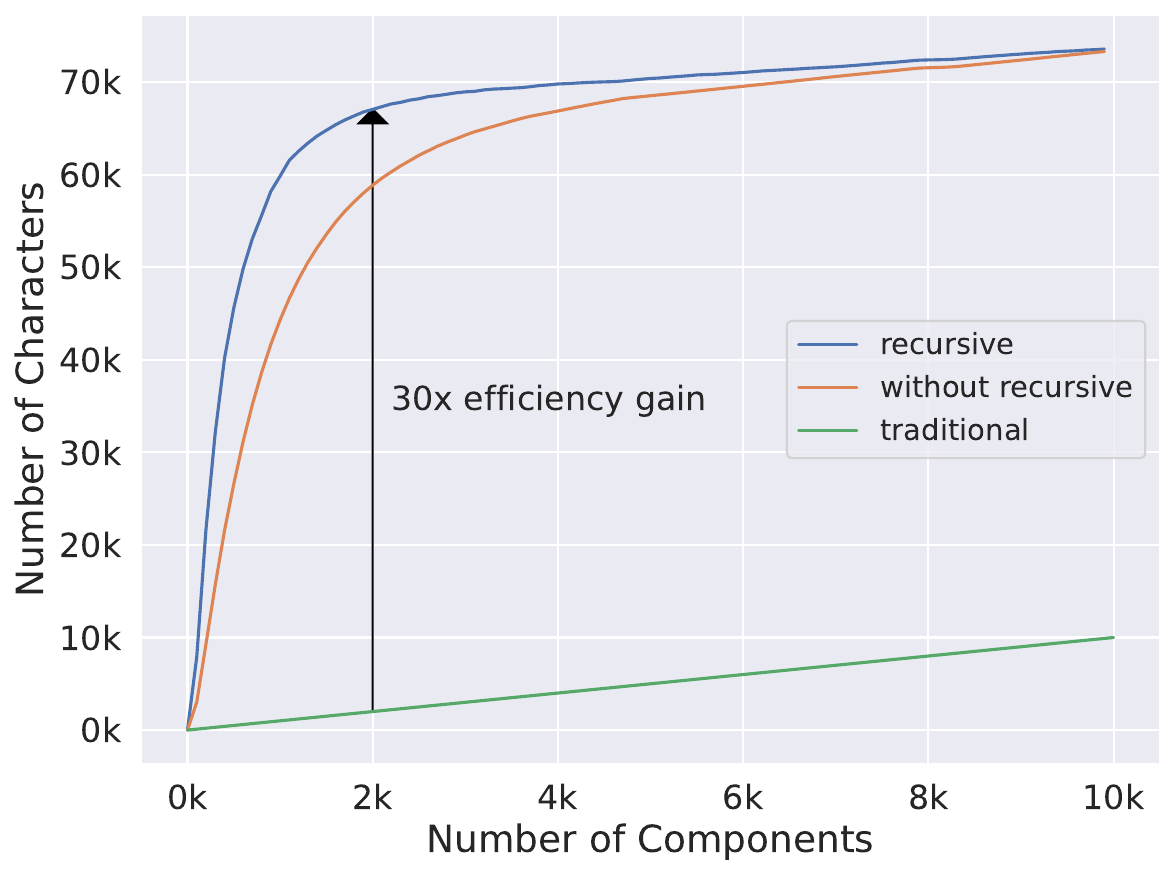}
    \caption{\textbf{\#components v.s. \#chars they could assemble.} Compared to traditional design pipeline, our approach is able to increase the efficiency by over 30 times.}
    \vspace{-1mm}
    \label{fig:design-speed}
\end{figure}

The layout distribution is illustrated in Fig.~\ref{fig:dist_layout}. Among the layouts, NL01 contains the most characters, while NL04 has the fewest. ``TBD" represents the unannotated data, which are usually omitted by designers and it is out of the scope of this paper. The distribution of variations in NL03 is depicted in Fig.~\ref{fig:sublayout}. These variations (row labels) are similar to each other, with slight differences. Fig.~\ref{fig:nested_dist} displays the number of nested characters within each layout. Nested characters refer to those composed of at least one composite component. Besides, we have also demonstrated the top 10 most frequently used components and the number of characters they belong to are shown in Fig.~\ref{fig:top_components}.

We also demonstrate the relationship between the number of components and the number of characters they can synthesize. The components are arranged in monotonic decreasing order \emph{w.r.t} their usage frequencies. As shown in Fig.~\ref{fig:design-speed}, the number of characters that the components could form increases rapidly.
The ``recursive" implies that the synthesize characters can be reused as new components to participate in the subsequent new character generation. Our approach can generate a font supporting over 60K characters with only 1.1K designed components, providing a huge boost to the traditional linear Chinese font design pipeline.

%% file: method.tex
\section{Component Affine Transformation Regressor}
The proposed Chinese font-components dataset allows us to synthesize characters by combining components using an affine transformation-based method. To be specific, we designed a neural network named CAR (\textbf{C}omponent \textbf{A}ffine transformation \textbf{R}egressor) to regress affine transformation matrix for each individual component image. 
Next, we apply the transformation matrix to its corresponding component to produce its transformed version with appropriate scale and position. Finally, these transformed components are merged to obtain the final synthetic character. Note that the model is trained in image space. At inference, the affine transformation can be directly applied to the vector font (\ie the control points). Compared to previous methods, our component composition-based method enables large-scale Chinese vector font generation in an efficient and scalable way. 

\subsection{Model Architecture}
\begin{figure*}
    \centering
    \includegraphics[width=0.98\textwidth]{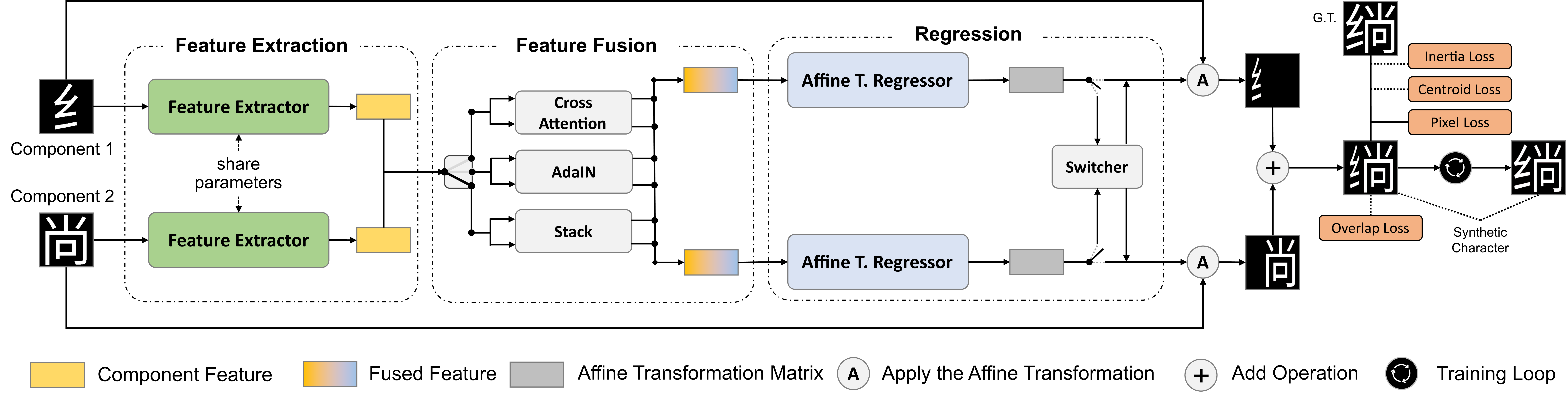}
    \vspace{-2mm}
    \caption{The overall architecture of our model.}
    \vspace{-2mm}
    \label{fig:model-arch}
\end{figure*}

For a character $C$ with components $R1$, $R2$, CAR produces synthetic character $S$ by merging transformed components. Our model resembles the Spatial Transformer Network~\cite{STN_NIPS2015_33ceb07b}, but we use it to regress component affine transformation matrices, whereas the original paper aims to enhance CNN representation ability via affine transformation. Fig.~\ref{fig:model-arch} shows the model architecture. 
The model consists of three modules: Feature extractor, fusion module, and the regressor.

\noindent\textbf{Feature extractor} is used to extract component features. It can be any backbone.
For simplicity, in our case, we utilized the classic VGG19-bn pretrained on ImageNet. The parameters are shared for different components.

\noindent\textbf{Feature fusion module} fuses features of components as input to the regressor.
The intuition behind the fusion module is that the regressor needs to know the information of all the components before it gives the appropriate transformation predictions for the current one. \comment{Take the left-right layout for example, if the right component has more strokes than the left one, the left one should be scaled down, and the vice versa.} For fusion strategy, Stack, AdaIN, and Cross Attention are all supported. The performance of these fusion strategies slightly differs \emph{w.r.t} different layouts. For more details, please refer to the supplementary material. 

\noindent\textbf{Regressor} takes as input the fused component features to compute the affine transformation matrices. The regressor is a $n$ layer MLP, with special initialization of the last linear layer. The weight is set to zero, while the bias term receives a standard affine transformation. This initialization ensures that our model performs identical transformations initially and gradually improves as training progresses. The resulting affine transformation matrices are applied to component images to obtain the transformed ones which are summed up to create the final character. 
\comment{We optimize the CAR with multiple objectives, whose details are illustrated in Sec.\ref{sec:loss}.}


\subsection{Loss} 
\label{sec:loss}
Our goal is to learn the reasonable affine transformation for each component and merge multiple components into a visually pleasing character. Since it is a difficult regression task, we designed multiple losses tailored to font characteristics to stabilize the training and penalize the bad cases. \comment{Although in inference stage, our method is used to generate vector characters, in training stage, the model is trained with images. }

First, the regular pixel-wise loss is computed to measure the difference between the synthesized character $S$ and ground truth $C$:
\begin{equation}
    L_{pixel} = \mathbf{D}(S, C)
\end{equation}
where $\mathbf{D}$ represents any distance function. In our case, we simply use L1 distance. 

The affine transformation matrix is initialized with an identity matrix such that the components may have a large overlap at the beginning of training loop. Therefore, the pixel loss $L_{pixel}$ contribute a lot in the beginning. However, when the networks converges, the overlap region becomes smaller and so as the pixel loss. As a consequence, the network stops pushing the two components away from each other.
To solve this issue and make each component move in the correct direction, we added the overlap loss:
\begin{equation}
    L_{overlap} = \frac{\sum_{x, y}\text{min}(\text{max}(S(x,y) {-} 1, 0), 1)}{\sum_{x,y}S(x,y)}
\end{equation}
where $S(x,y)$ denotes the pixel value of $S$. In the two-radical synthesis case, the pixel value that is greater than 1 means two radicals overlap there. 

Besides, the centroid is another important attribute of Chinese font characters. The reasonable synthesized character $S$ should share the same centroid position as the ground truth $C$ to guarantee the balance of synthesized character. Therefore, we also introduced a centroid loss:
\begin{equation}
\begin{split}
    L_{cent.} = &\frac{1}{2} (\mathbf{D}(\frac{\Phi(S, 1, 0)}{\Phi(S, 0, 0)}, \frac{\Phi(C, 1, 0)}{\Phi(C, 0, 0)}) \\ &+ \mathbf{D}(\frac{\Phi(S, 0, 1)}{\Phi(S, 0, 0)}, \frac{\Phi(C, 0, 1)}{\Phi(C, 0, 0)}) )
\end{split}
\end{equation}
where $\Phi$ represents raw moments:
\begin{equation}
\Phi(I, i, j) = \sum_{x,y} {I(x,y) x^i y^j}
\end{equation}

To make sure that every pixel of synthesized character $S$ has the same distance to the centroid as the ground truth $C$, we introduced second order central moment, \emph{i.e.} inertia loss.
\begin{equation}
\begin{split}
    L_{inertia} &= \mathbf{D}(\Psi(S), \Psi(C))
\end{split}
\end{equation}
where the $\Psi$ represents the inertia that is calculated as follows:
\begin{equation}
    \Psi(I) = \Phi(I, 2, 0) {-} \frac{\Phi(I, 1, 0)^2}{\Phi(I, 0, 0)} 
        {+} \Phi(I, 0, 2) {-} \frac{\Phi(I, 0, 1)^2}{\Phi(I, 0, 0)}
\end{equation}
In practice, the inertia loss regulated the shape and bounding box of the synthesized character.


The final loss function is then defined as:
\begin{equation}
\label{eq:loss}
\begin{split}
    \mathcal{L} & = \alpha L_{pixel} + \beta L_{overlap} + \gamma L_{cent.} + \theta L_{inertia}
\end{split}
\end{equation}

The lowercase Greek letters denote the weight of the loss. 


\subsection{Implementation Details for Different Layouts}
\label{sec:details_for_layouts}
As illustrated in Sec.\ref{sec:comp_and_layout}, we divide the Chinese font into six layouts. For all layouts, we centralize the component image, randomly flip them, and feed them into the CAR for training. \comment{For NL01 (Left-Right), we inited the extractor with pretrained parameters and trained the CAR from scratch. For NL02(Top-Bottom), we proposed a switcher module to make the NL01-pretrained affine transformation adapt to NL02 characters, allowing NL01 pretrained model to be continued training on NL02 characters. For NL03 (Enclosed), we increase the weight of centroid and inertia to make CAR converge. For NL04 (Left-Middle-Right) and NL05 (Top-Middle-Bottom), we proposed an iterative invocation strategy to improve the performance. Due to page constraints, more information is provided in the supplementary material.}


\begin{enumerate}[leftmargin=0.4cm]
\item For NL01 (Left-Right), the CAR is trained from scratch, but the feature extractor is initialized with the pretrained parameters. 
\item For NL02 (Top-Bottom), we first attempted to train the CAR from scratch. To improve the performance, the intuitive idea is to use the parameters of NL01 pre-trained models. Since the NL01 characters are quite different from NL02's, the affine transformation regressor trained on NL01 is not directly suitble for NL02. To overcome this, we introduced a switcher module that\comment{ is a $6\times6$ diagonal matrix with all positions set to 0 except for (0,4), (4,0), (2,5), and (5,2), which have values of 1. This module} swaps the parameters affecting the horizontal and vertical axes in the affine transformation, allowing the NL01 affine transformation matrices to adapt to NL02 characters. With the pretrained parameters, the CAR of NL02 converges faster and gain better performance. 
\item For NL03 (Enclosed), the base CAR failed easily. To solve this, we increased the weight of centroid and inertia loss to make the CAR converge. 
\item For NL04 (Left-Middle-Right) and NL05 (Top-Middle-Bottom), we adjusted our model to this three-components font synthesis task. To further improve the performance, we proposed an iterative invocation strategy where we used pretrained NL01/NL02 CAR to perform the NL04/NL05 synthesis task. For example, for an NL04 character, we first invoke the NL01 pretrained CAR to synthesize the intermediate character with the first two components. We then invoke the NL01 pretrained CAR again to synthesize the final character based on the intermediate character and the last component. Experimental results demonstrate the effectiveness of our method. More details are provided in the supplementary materials.
\end{enumerate}




\subsection{Integration with Glyphs}
\label{subsec: glyphs}
Our method targets Chinese font designers and their design process. We have written a plug-in for Glyphs, a widely-used font design tool, to enable designers to rapidly generate numerous style-consistent, editable characters based on carefully designed components. The details and pseudocode are available in the supplementary material.

%% file: evaluation.tex
\section{Experiments}

\subsection{Experimental Settings}
\label{sec:experiment_setup}
To evaluate our method, we collected seven fonts mainly used in body text, including SourceHanSans, BableStone, AlibabaPuHuiTi, OPPOSansR, SmileySans, YouAiYuanTi, and WenDingKaiTi. We rendered the characters supported by the font along with their components into images and categorized these images according to the layout type of the characters to serve as evaluation data for each layout. We used 80\% of the character images from one or more fonts as training data and the remaining 20\% as validation data. We adopted multiple metrics for quantitative evaluation, including MAE, RMSE, FID~\cite{heusel2017gans}, and LPIPS~\cite{zhang2018unreasonable}. MAE and RMSE describe the reconstruction error at the pixel level. LPIPS measures perceptual similarity between synthesized characters and ground truth. FID measures the distance between the target and generation distribution. Although our method is not a typical generative model but a regression model for synthesizing characters with regressed affine transformation matrices, we still report FID because we believe it can accurately represent the synthesis quality. Please refer to the supplementary material for more details on the experiments.

\subsection{Ablation Study}

\begin{figure}
    \centering
    \includegraphics[width=0.95\linewidth]{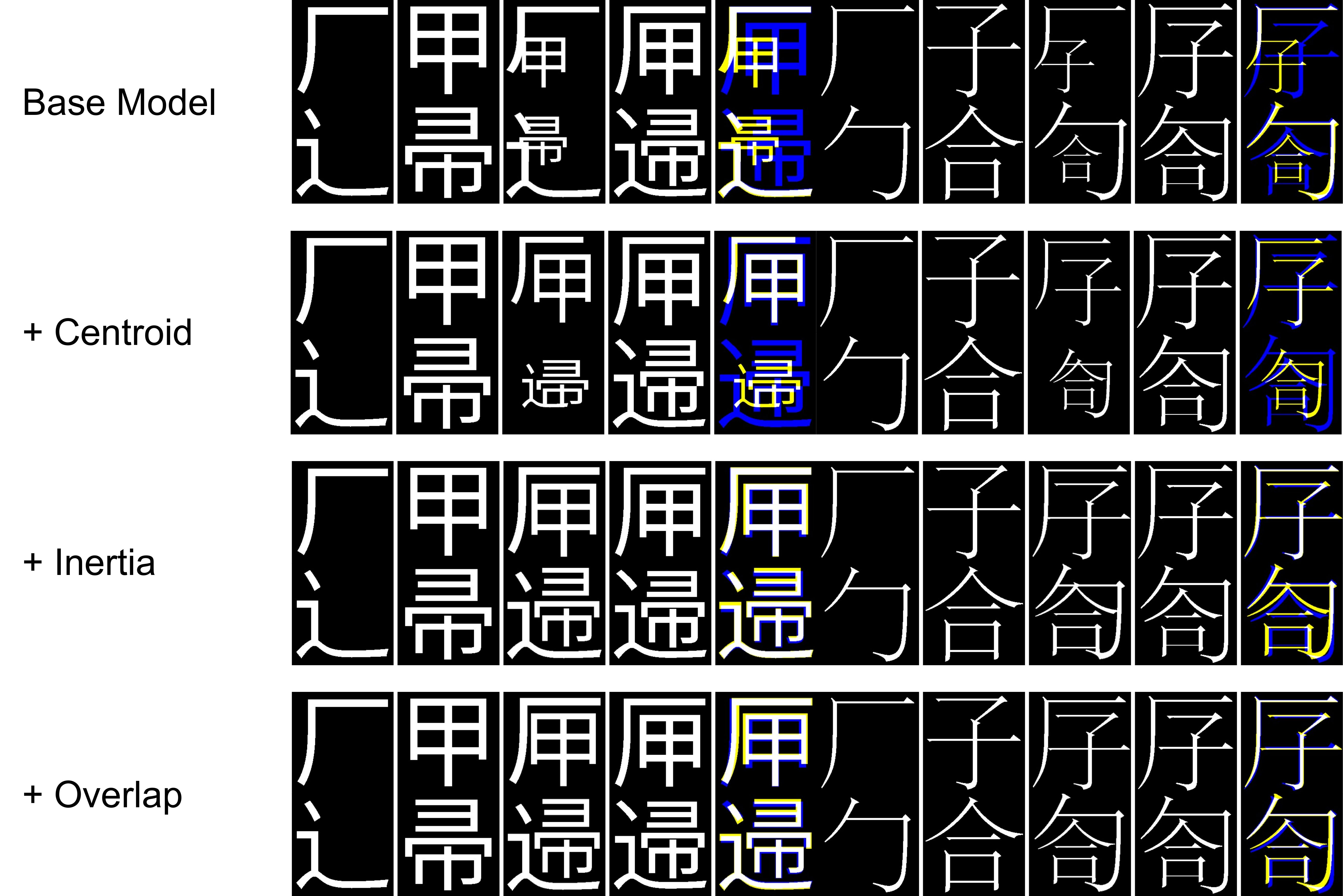}
    \vspace{-1mm}
    \caption{\textbf{Ablation study of losses.} \normalfont{From left to right, the images represent component 1, component 2, synthetic character, ground truth, and the difference between synthetic character and ground truth, respectively. \jy{The color shows the difference between the GT (blue) and the synthetic (yellow) characters. Their overlap is white.}}}
    \label{fig:ablation_study_losses}
    \vspace{-2mm}
\end{figure}

\noindent\textbf{Effectiveness of the Losses.} We conducted experiments to evaluate the effectiveness of losses. First, we enabled only pixel loss to establish a baseline. Then, we successively enabled centroid, inertia, and overlap losses for comparison. CARs used in the experiments were all trained on SourceHanSans in NL03. Tab.~\ref{tab:losses} and Fig.~\ref{fig:ablation_study_losses} show the quantitative results and generated samples, respectively. The results indicate that as these losses were gradually added, the performance improved continuously.

\begin{table}[ht]
\small
\centering
\begin{tabular}{l|cccc}
\hline
Model Setting      & \multicolumn{1}{c}{MAE↓} & \multicolumn{1}{c}{RMSE↓} & \multicolumn{1}{c}{FID↓} & \multicolumn{1}{c}{LPIPS↓} \\ \hline
Base Model & 0.2064                  & 0.4499                   & 81.8785                 & 0.3799                    \\
+ Centroid & 0.1912                  & 0.4309                   & 49.5050                 & 0.3150                    \\
+ Inertia  & 0.1936                  & 0.4316                   & 35.8110                 & 0.2757                    \\
+ Overlap  & \textbf{0.1891}                 & \textbf{0.4264}                   & \textbf{33.4290}                 & \textbf{0.2709}                    \\ \hline
\end{tabular}
\caption{\textbf{Ablation study of losses.}}
\vspace{-1mm}
\label{tab:losses}
\end{table}

\noindent\textbf{Effectiveness of the Feature Fusion.} To evaluate the feature fusion module, we compared the performance of CARs trained on SourceHanSans, both with and without feature fusion. Tab.\ref{tab:fusion} shows that the model with feature fusion consistently improved overall performance, demonstrating the module's effectiveness. Without feature fusion, the regressor tended to generate a constant affine transformation for specific components, resulting in performance degradation.

\begin{table}[ht]
\small
\begin{tabular}{ll|cccc}
\hline
Layout & Fusion         & MAE↓            & RMSE↓             & FID↓              & LPIPS↓ \\ \hline
NL01   & $ \checkmark $     & \textbf{0.1312} & \textbf{0.3540}   &  \textbf{12.7156} & \textbf{0.2019} \\
NL01   & ×              & 0.1676          & 0.4022            & 13.6127           & 0.2332 \\ \hline
NL02   & $ \checkmark $     & \textbf{0.1562} & \textbf{0.3880}   &  \textbf{13.0982} & \textbf{0.2210} \\ 
NL02   & ×              & 0.2630          & 0.5079            & 17.8102           & 0.3234 \\ \hline
NL03$_{0,1}$   & $ \checkmark $     & \textbf{0.1891} & \textbf{0.4264}   & \textbf{33.4290}  & \textbf{0.2709} \\
NL03$_{0,1}$  & ×              & 0.2687          & 0.2615            & 59.3720         & 0.4039 \\ 
\hline
\end{tabular}
\caption{\textbf{Results of the feature fusion.} NL03$_{0,1}$ denote the models are trained on the data of the first and second variations of NL03.}
\vspace{-1mm}
\label{tab:fusion}
\end{table}


\noindent\textbf{Effectiveness of the Switcher Module.} We compared the performance of CAR with and without switcher module on SourceHanSans NL02. \comment{two implementations of NL02 described in Section~\ref{sec:details_for_layouts} to demonstrate the effectiveness of the current implementation. The first implementation involved directly training an NL02 model from scratch. In the second implementation, we used a switcher module to rotate the affine transformation, allowing us to continue training the CAR on NL02 data based on a NL01 pretrained model.} Tab.~\ref{tab:nl02-comparison} presents the quantitative results and illustrates the effectiveness.

\begin{table}[ht]
\centering
\small
\begin{tabularx}{\linewidth}{Xc|cccc}
\hline
Layout & Switcher & MAE↓             & RMSE↓            & FID↓              & LPIPS↓           \\ \hline
NL02 & ×  & 0.1683          & 0.4035          & 13.2593          & 0.2341          \\
NL02 & $ \checkmark $  & \textbf{0.1562} & \textbf{0.3880} & \textbf{13.0982} & \textbf{0.2210} \\ 
\hline
\end{tabularx}
\caption{\textbf{Ablation study of switcher.}}
\vspace{-1mm}
\label{tab:nl02-comparison}
\end{table}


\noindent\textbf{Effectiveness of the Iterative Invocation Strategy.} We compared our implementation using iterative invocation strategy with a three-components CAR trained from scratch. The CARs used in this comparison were trained on SourceHanSans and BableStone data, as the available characters for NL04 and NL05 are too limited. As shown in Tab.~\ref{tab:iter_invoc}, the results confirms the effectiveness of the strategy. 

\begin{table}[ht]
\small
\begin{tabular}{ll|cccc}
\hline
Model              & Layout & MAE↓            & RMSE↓           & FID↓             & LPIPS↓          \\ \hline
\textbf{NL01 I.I.}  & NL04   & \textbf{0.2270} & \textbf{0.4683} & \textbf{67.0076} & \textbf{0.3407} \\
NL04               & NL04   & 0.2474          & 0.4913          & 97.0602          & 0.3782          \\ \hline
\textbf{NL02 I.I.} & NL05   & \textbf{0.1836} & \textbf{0.4212} & \textbf{33.0506} & \textbf{0.3209} \\
NL05               & NL05   & 0.2217          & 0.4648          & 54.8968          & 0.3815          \\ \hline
\end{tabular}
\caption{\textbf{Ablation study of iterative invocation strategy.} \normalfont{I.I. represents iterative invocation.}}
\vspace{-1mm}
\label{tab:iter_invoc}
\end{table}

\subsection{Font Generation}
\label{sec:font_generation}

Our method allows for the composition of different components to generate Chinese characters including complex ones. Regarding comparison, we found it inappropriate to compare our component composition-based method with other style transfer-based methods. Therefore, we first implemented three types of component composition-based GANs as baselines. Nevertheless, we compared our method with the SOTA Chinese vector font generation methods, DVF-v2~\cite{wang2023deepvecfont}, in generating complex Chinese characters. 
The details of comparison and implementation are provided in the supplementary material. \jy{Note that DVF~\cite{wang2021deepvecfont} is not included in the comparison because it failed to converge when being trained on complex Chinese characters. Non-Chinese vector font generation methods (~\eg DeepSVG) are also not included because previous work~\cite{wang2021deepvecfont,aoki2022svg,wang2023deepvecfont} has proved their shortcomings in generating Chinese vector characters.}

\begin{figure}[ht]
    \centering
    \includegraphics[width=\linewidth]{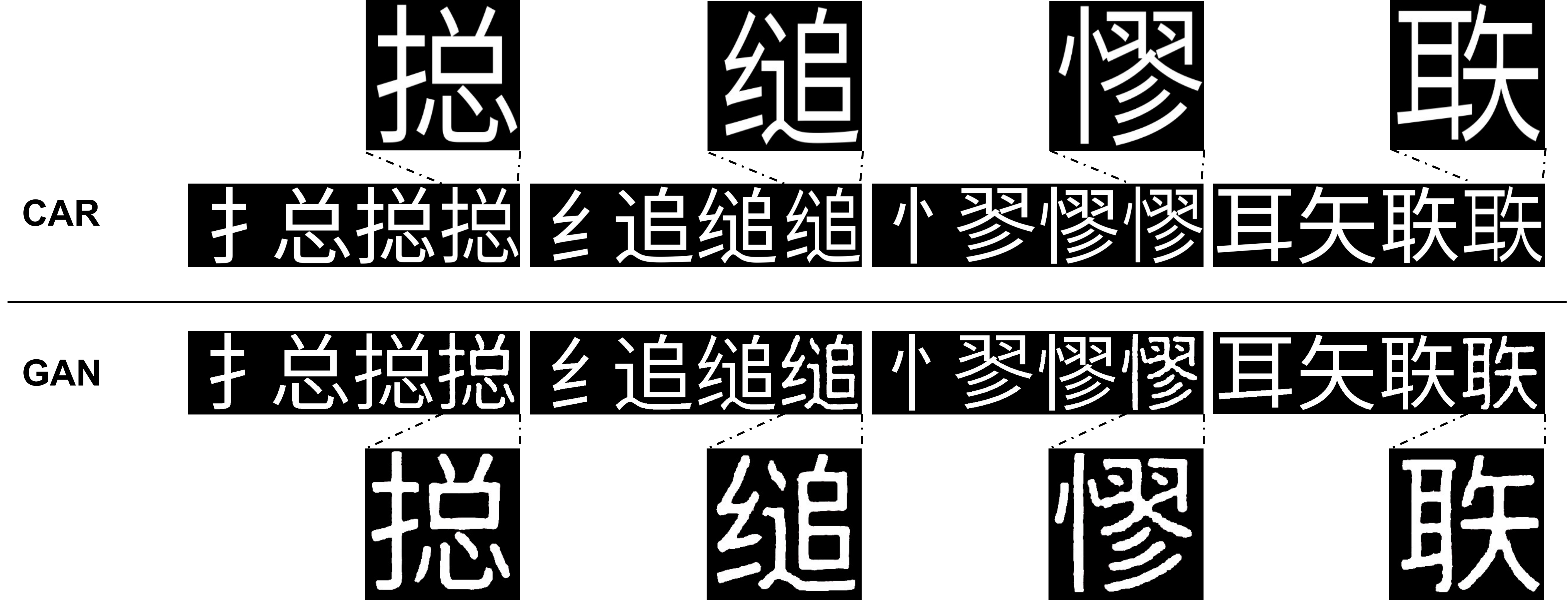}
    \caption{\textbf{Generated samples of our CARs and GANs.} \normalfont{For each group, from left to right, the figure displays the component 1, component 2, ground truth, and synthetic character, respectively.}}
    \label{fig:font_gen_results}
    \vspace{-2mm}
\end{figure}

\begin{figure}[ht]
    \centering
    \includegraphics[width=\linewidth]{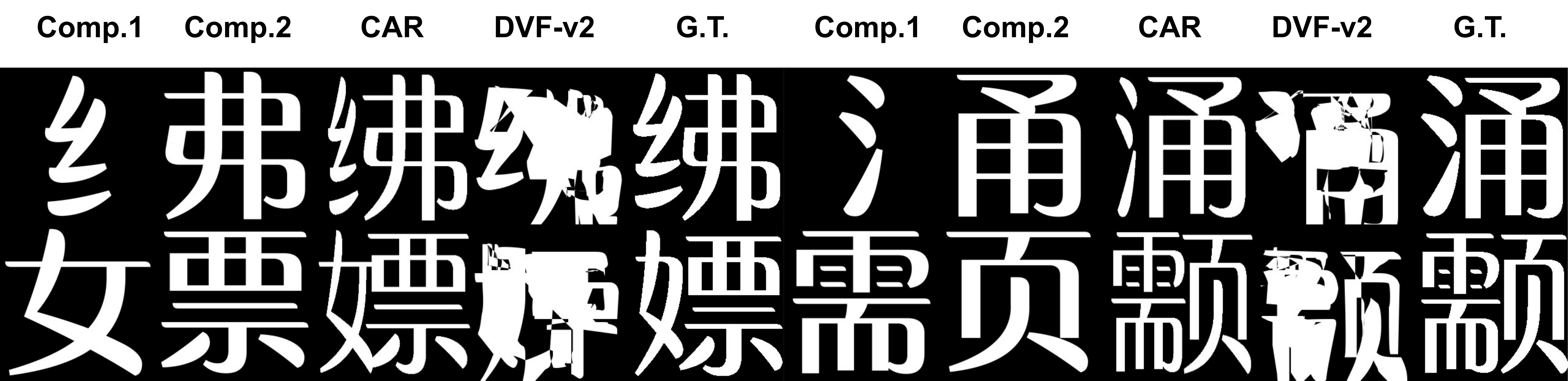}
    \caption{Generated results of our CAR and DVF-v2. }
    \vspace{-1mm}
    \label{fig:dvf_car_results}
\end{figure}

Not surprisingly, our method outperforms GANs and DVF-v2 across all metrics, as evidenced in Tab.~\ref{tab:font_generation} and ~\ref{tab:complex_comparison}. Specifically, our CAR achieves an improvement in MAE by \textit{+0.05} over GAN-\uppercase\expandafter{\romannumeral2} and \textit{+0.206} over DVF-v2, and in LPIPS by \textit{+0.134} over GAN-\uppercase\expandafter{\romannumeral2} and \textit{+0.223} over DVF-v2. This highlights its exceptional generative abilities, especially for complex characters. As shown in Fig.~\ref{fig:font_gen_results} and ~\ref{fig:dvf_car_results}, while GANs produce blurry outputs and DVF-v2 struggles with complex characters, our method synthesizes these complex characters in much better quality, mainly due to the introduction of affine transformation-based component composition pipeline. More generated examples are provided in supplementary material.

\begin{table}[ht]
\small
\centering
\begin{tabularx}{\linewidth}{ll|XXcX}
\hline
Model    & Layout      & MAE↓    & RMSE↓   & FID↓      & LPIPS↓  \\ \hline

CAR      & NL02         & 0.1562 & 0.3880 & 13.10  & 0.2210 \\
CAR*   & NL03$_{0,1}$     & 0.1891 & 0.4264 & 33.43  & 0.2709 \\
CAR*   & NL03$_{2,3,4,5}$ & 0.1446 & 0.3714 & 35.30  & 0.2140 \\
CAR*   & NL03$_6$       & 0.2524 & 0.4947 & 45.86  & 0.3911 \\
CAR I.I. & NL04         & 0.2270 & 0.4683 & 67.01  & 0.3407 \\
CAR I.I.  & NL05         & 0.1836 & 0.4212 & 33.05  & 0.3209 \\ \hline
CAR      & NL01         & \textbf{0.1312} & \textbf{0.3540} & \textbf{12.72}  & \textbf{0.2019} \\
\hline
GAN-\uppercase\expandafter{\romannumeral1}    & \multicolumn{1}{l|}{NL01}         & 0.2649 & 0.7064 & 178.71 & 0.3472 \\
GAN-\uppercase\expandafter{\romannumeral2}    & \multicolumn{1}{l|}{NL01}         & 0.1851 & 0.5987 & 151.32 & 0.3359 \\
GAN-\uppercase\expandafter{\romannumeral3}    & \multicolumn{1}{l|}{NL01}         & 0.2203 & 0.6539 & 172.97 & 0.3632 \\ 
\hline
\end{tabularx}
\caption{\textbf{Results of font generation}. The CAR* represents the CAR model with all losses enabled. \comment{The NL03$_{x}$ represents the CAR trained on characters of specific variations of NL03. For example, NL03$_{0}$ means the first variation of NL03.}  \\}
\label{tab:font_generation}

\begin{tabular}{l|cccc}
\hline
Model  & MAE↓    & RMSE↓   & FID↓      & LPIPS↓  \\ \hline
DVF-v2 & 0.3777 & 0.6042 & 283.34 & 0.4671 \\ 
\textbf{CAR (Ours)} & \textbf{0.1713}	& \textbf{0.4068} & \textbf{48.21} & \textbf{0.2446} \\ \hline
\end{tabular}
\caption{\textbf{Results of complex Chinese characters generation} compared with SOTA DVF-v2.} 
\vspace{-1mm}
\label{tab:complex_comparison}

\end{table}


\comment{Once again, as a regression model, it may not be entirely appropriate to use the generation task metrics for evaluation, but here we only use it to describe the generation capability of our method, which is the prerequisite for assisting font design. }


To further evaluate the font generation ability, we conducted a user study to see if the fonts we generated were good enough for users. 
There were 55 participants in this study and the results are shown in Tab.~\ref{tab:font-generation-eval}. 

\begin{table}[!ht]
\small
\centering
\begin{tabular}{l|lll}
\hline
Character Type & GAN-gen & CAR-gen & Designer \\ \hline
Win Rate       & 25.45\% & 54.55\% & 76.36\% \\
\hline
\end{tabular}
\caption{\textbf{Results of the qualitative evaluation of font generation.} GAN-gen, CAR-gen, and Designer denotes the creator of character. The win rate represents what percentage of people think the image is not generated by algorithm.}
\vspace{-2mm}
\label{tab:font-generation-eval}
\end{table}


\subsection{Zero-Shot Font Extension} 
\begin{figure}[!htb]
    \centering
    \includegraphics[width=\linewidth]{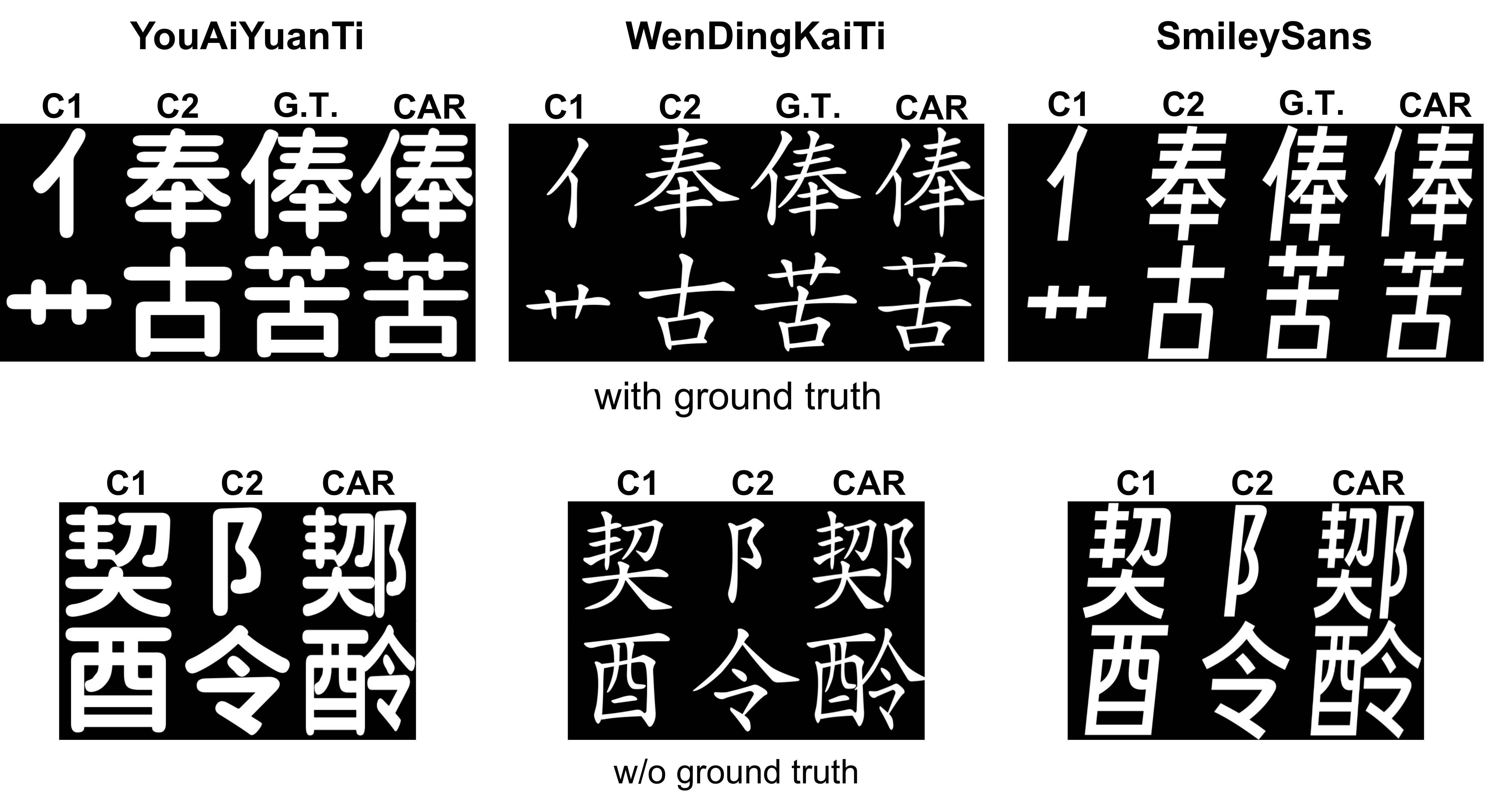}  
    \caption{\textbf{Zero-shot extension results} of YouAiYuanTi, WenDinGKaiTi and SmileySans. \comment{For the upper half, each column represents component 1, component 2, ground truth, and generated character, respectively. For the bottom, each column represents component 1, component 2, and synthetic character, respectively.}}
    \label{fig:zero-shot-generation}
    \vspace{-2mm}
\end{figure}

Our vector font generation method enables it to act as an out-of-the-box font extension tool for multiple fonts with varying styles in a zero-shot manner. Firstly, given a released font set, we extract the supported components. We then use our model to extend the released font set. 
The zero-shot font generation results on YouAiYuanTi, WenDingKaiTi, and SmileySans are shown in Fig.~\ref{fig:zero-shot-generation}, demonstrating the effectiveness of our method in the zero-shot font extension scenario.


We further conducted a user study with 90 participants to verify the improvement on user experience with our font completion. We provide the details of the user study in our supplementary material. We analyzed their preferences between our completed version and the original version. The results shown in Tab.~\ref{tab:font-completion} demonstrate the effectiveness of our method in improving user experience.

\begin{table}[!htb]
\small
\centering
\begin{tabular}{l|cc}
\hline
Font       & Completed & Orginal \\ \hline
Preference & 90\%   & 10\% \\ \hline
\end{tabular}
\caption{The text rendered by \jy{the font extended by our method} is preferred by the most individuals.}
\vspace{-2mm}
\label{tab:font-completion}
\end{table}

%% file: conclusion.tex
\section{Conclusion}

\comment{In this paper, we present a large-scale Chinese font-components dataset that covers over 90K Chinese characters defined in Unicode. Building on this dataset, we propose an efficient and scalable Chinese font generation method that uses neural networks to learn affine transformations. The learned transformations can be applied directly to the control points of vector fonts to synthesize characters in vector format. Additionally, we developed a Glyphs plugin to integrate our method into the Chinese font design pipeline to truly boost the Chinese font design. What's more, our work not only can expand the released Chinese font sets to improve user experience, but also has the potential to generate rarely-used characters and contribute to the preservation of Chinese culture. To further advance our work, future research will focus on optimizing the dataset by improving the labeling accuracy, and enhancing the quality of details for the generated Chinese fonts.}

In this paper, we first present the largest Chinese font-components dataset that covers over 90K Chinese characters. Based on this dataset, we propose an simple yet effective Chinese vector font generation framework where the character generation task is treated as a component composition task. Compared to style transfer-based methods, our method introduces a new pipeline to generate large-scale Chinese vector font via component composition. Additionally, we develop a Glyphs plugin to integrate our method into the Chinese font design pipeline to truly boost Chinese font design. Extensive experiments evidence that our method surpasses the SOTA vector font generation methods in generating complex Chinese characters in both new font generation and zero-shot font extension. Furthermore, our future work will focus on refining the annotations in our dataset, and enhancing the quality of the generated Chinese characters.

%% file: supp-comparison.tex
\section{Comparison with Vector-based Methods}

\jy{Tab.~\ref{tab:comp-vec-methods} shows a detailed comparison with current vector-based methods. We'd like to emphasize \textit{four} differences between prior works and ours, which indicates the novelty and contribution of our work: 1) \textbf{Task \& Capability}: Previous works generate \textit{simple} vector graphics (e.g. alphabets, simple Chinese characters), while our method synthesizes \textit{complex} Chinese characters with \textit{high} quality that significant surpasses previous SOTA. 2) \textbf{Paradigm}: Previous vector-based Chinese font generation methods (abbr. PVC) are mainly based on \textit{style transfer} and trained under \textit{vector (and image)} supervision, while ours is based on \textit{component composition} and trained with only \textit{image} supervision, streamlining the training and generation process. 3) \textbf{Generalization}: PVCs only support generating \textit{fixed} and \textit{limited} number of Chinese characters \textit{seen} in training, 
whereas our method is capable of generating characters \textit{unseen} during the training and can generate over \textit{75,000} Chinese characters given sufficient components (Fig. 5). 4) \textbf{Practical Implications}: PVCs are \textit{barely applicable} in real-world scenarios due to mentioned limitations, while ours is \textit{able} to accelerate the real Chinese character design process.}


\begin{table*}[ht]
\centering
\small
\begin{tabular}{@{}lcccccccc@{}}
\toprule
\multirow{2}{*}{Methods} & \multicolumn{3}{c}{Capability}        & \multicolumn{3}{c}{Generalization} & \multicolumn{2}{c}{Feature}                       \\ \cmidrule(l){2-9} 
                         & Chinese & Complexity & Control Points & Quantity    & Unseen   & Quality   & Paradigm                        & Supervision     \\ \midrule
SVG-VAE & ✗ & -       & -        & -            & - & -    & -                     & Vector          \\
DeepSVG & ✗ & -       & -        & -            & - & -    & -                     & Vector          \\
Im2Vec  & ✗ & -       & -        & -            & - & -    & -                     & Image           \\
DVF                      & ✓       & Simple     & $\sim$50       & Limited     & ✗        & Lowest    & \multirow{2}{*}{Style Transfer} & Image \& Vector \\
DVF-v2  & ✓ & Simple  & $\sim$70 & Limited      & ✗ & Low  &                       & Image \& Vector \\ \midrule
Ours    & ✓ & Complex & $\infty$        & 75,000+ & ✓ & High & Component Composition & Image           \\ \bottomrule
\end{tabular}
\caption{Comparison with Vector-based Methods}
\label{tab:comp-vec-methods}
\end{table*}

%% file: supp-dataset.tex
\section{Dataset}
\subsection{Radical and Component}
The majority of Chinese characters are composed of different components. To elucidate the notion of a component, we initially introduce the concept of Chinese character radicals. According to the \textit{Table of Indexing Chinese Character Components}~\footnote{\url{https://en.wikipedia.org/wiki/Table_of_Indexing_Chinese_Character_Components}}, there are 301 standard radicals.

\begin{figure}[ht]
    \centering
    \includegraphics[width=\linewidth]{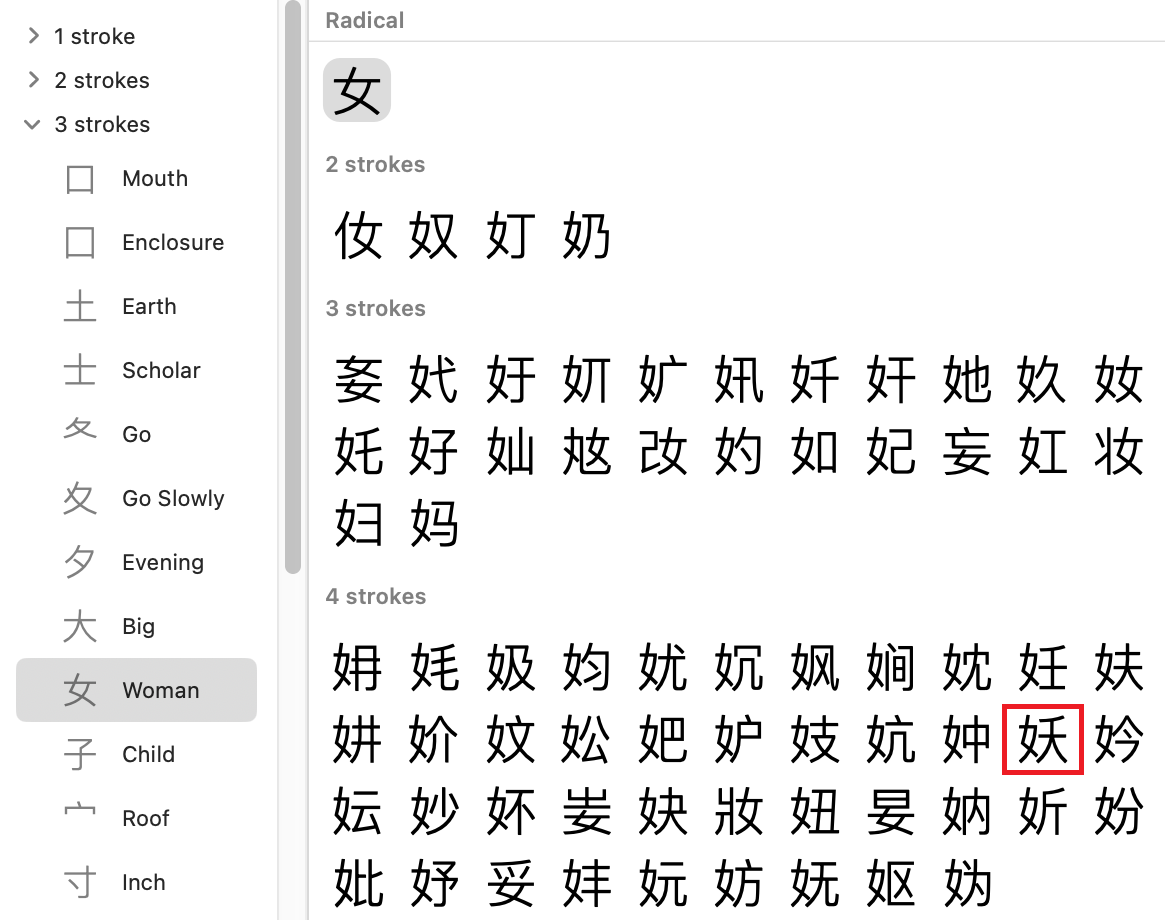}
    \caption{Chinese characters composed by \cnw{女} with different strokes.}
    \label{fig:radical}
\end{figure}

\jy{Figure~\ref{fig:radical} illustrates the radical system and the characters composed by the radical~\cnw{女}. However, radicals are primarily employed for Chinese character retrieval, as they pertain to only one portion of the characters. For instance, in Figure~\ref{fig:radical}, ~\cnw{夭}, the right part of \cnw{妖} (\ie the character in the red box), is not considered a radical. In other words, radicals do not represent every element of a separable character upon division. Consequently, we adopt another concept, component, to signify each part of a separable character. For example, both \cnw{女} (\ie left part) and \cnw{夭} (\ie right part) are considered components of \cnw{妖}.}

Evidently, the number of components significantly exceeds that of radicals. In our dataset, excluding nested definitions, there are approximately 11K components.

\subsection{Construction of our Dataset}
To annotate the required layout and component information within the dataset, we employed a crowdsourcing approach due to the dataset's extensive scale. In collaboration with a Chinese crowdsourcing company, we partitioned the dataset into multiple segments based on character ID, assigning each segment to crowdsourcing workers for layout and component annotation. Prior to launching the crowdsourcing efforts, we conducted standardized training, offering clear explanations to the workers regarding dataset information, required annotations, and criteria for decomposing radicals and components.

Throughout the annotation process, we carried out periodic random checks on the results and requested improvements from the workers as needed. Upon the annotation's completion, we enlisted 15 volunteer graduate students to review the annotated outcomes, thereby ensuring annotation quality. The review process entailed two rounds, with each volunteer assigned different content; we also performed regular random checks on the review results and implemented improvements when necessary.

\subsection{Accessibility}
We provide a minimal version of our dataset with the Anonymous Github link~\footnote{\url{https://anonymous.4open.science/r/Hanzi-database-subset-DB01/}} for review purposes. Full access to the dataset is possible upon a request and authorization process.

\begin{figure*}[htb]
    \centering
    \includegraphics[width=\textwidth]{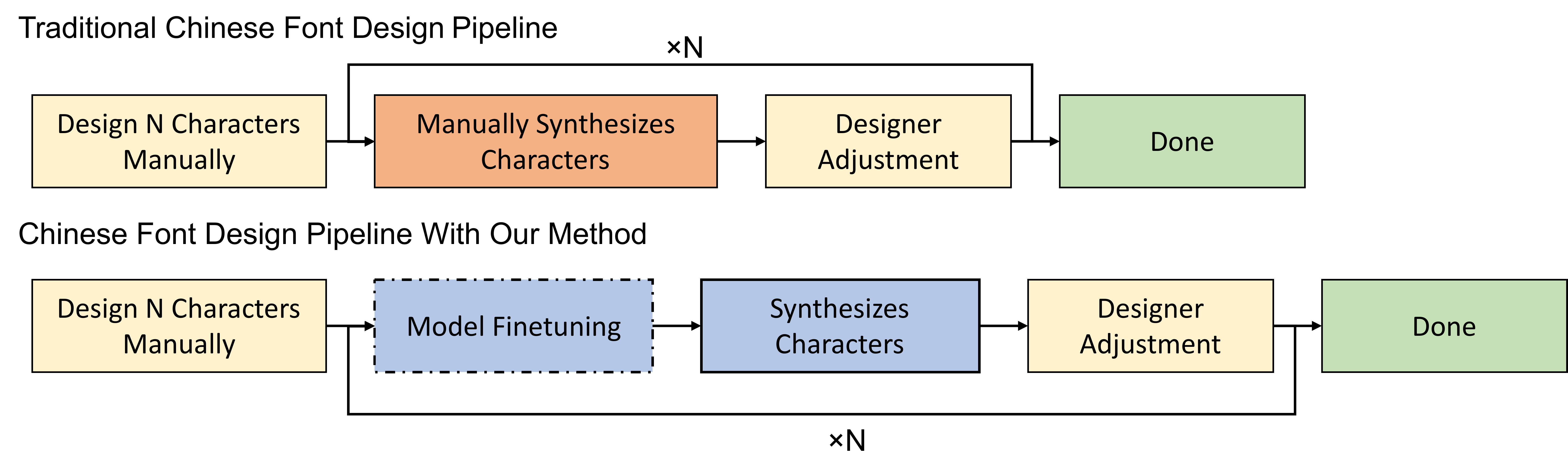}
    \caption{The paradigm shift of Chinese font design with our method. }
    \label{fig:paradigm_shift}
\end{figure*}

%% file: supp-method.tex
\section{Method}
\subsection{Training Visualization}
\begin{figure}[htb]
    \centering
    \includegraphics[width=\linewidth]{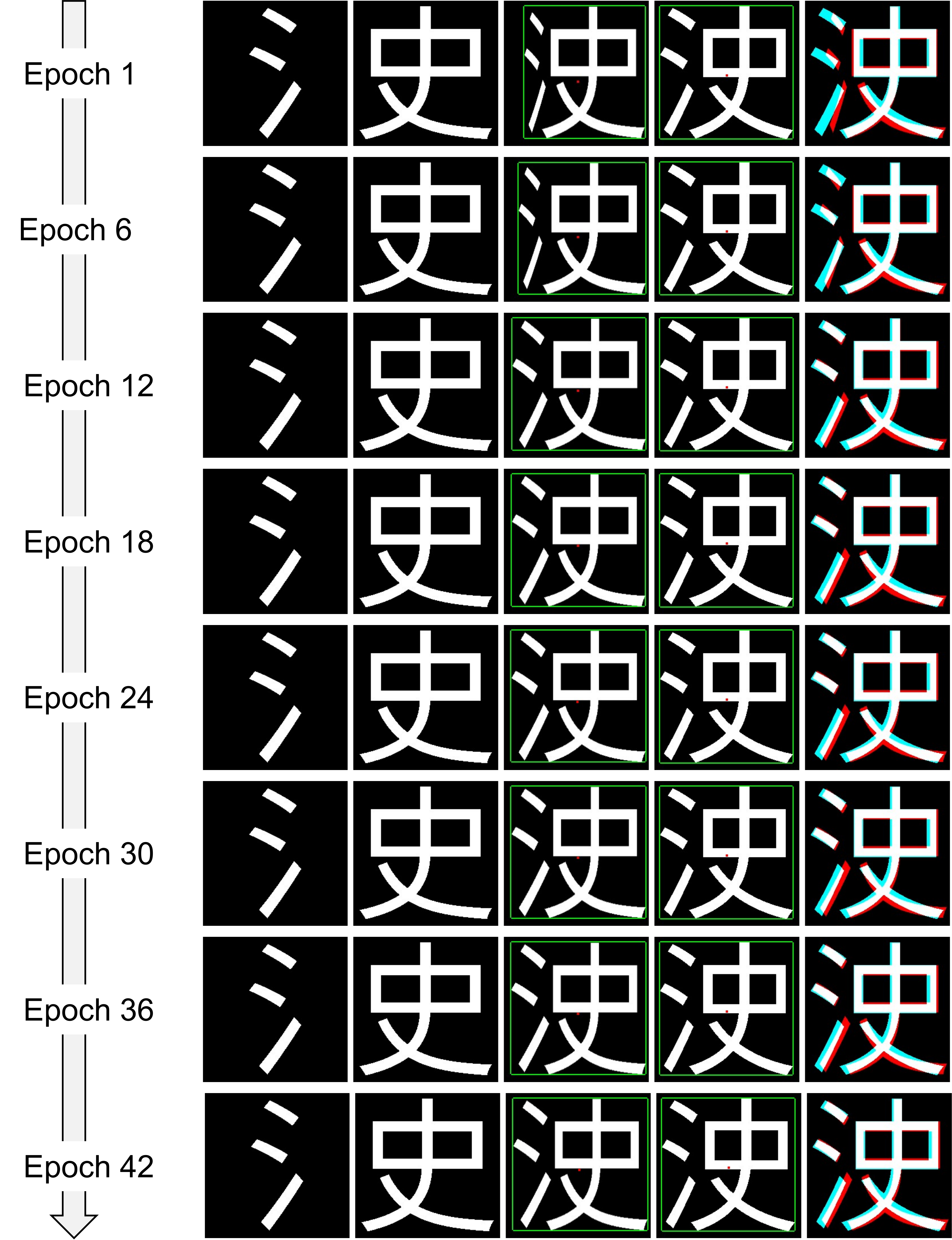}
    \caption{The visualization of training process. \normalfont{For each row, from left to right, there are component 1, component 2, generated character, ground truth, and difference between generated character and ground truth, respectively.}}
    \label{fig:training-visualization}
\end{figure}

We show the training process of our model in Figure~\ref{fig:training-visualization}. As the training progresses, our method exhibits an increasingly refined synthesis of characters, approaching a higher level of fidelity to the ground truth.

\subsection{Fusion Strategies}
We implemented three types of fusion method, namely Stack, AdaIN, and Cross Attention. As for Stack, we simply concatenate them together in channel dimension. As for AdaIN, we compute feature using the equation below:
\begin{equation}
\label{eq:adain}
    \hat{f_{R1}} = \sigma(f_{R2})\frac{f_{R1} - \mu(f_{R1})}{\sigma(f_{R1})} + \mu(f_{R2})
\end{equation}
where $\hat{f_{x}}$ refers to the fused feature, $f_{x}$ represents the feature of components $x$, and $\mu$ and $\sigma$ represents the mean and std of the feature, respectively. For cross attention fusion, the fused feature is calculated using the following equation:
\begin{equation}
\label{eq:attention}
    \hat{f_{R1}} = Softmax(\frac{Q(f_{R1})K(f_{R2})^T}{\sqrt{d}}) V(f_{R2})
\end{equation}
where $Q$, $K$, and $V$ represent the linear layers which output query, key, and value vectors for attention, respectively.

We additionally tested the performance of \sg{our proposed} CAR using SPADE~\cite{park2019semantic} as a feature fusion method. The affine parameters of SPADE are learned from feature of another component. 


\begin{table}[htb]
\begin{tabularx}{\linewidth}{ll|XXXX}
\hline
Layout & Fusion    & MAE↓   & RSME↓  & FID↓    & LPIPS↓ \\ \hline
NL01   & Stack     & \textbf{0.1312} & \textbf{0.3540} & 12.7156 & \textbf{0.2019} \\
NL01   & AdaIN     & 0.1385 & 0.3641 & \textbf{12.6574} & 0.2085 \\
NL01   & Attention & 0.1426 & 0.3699 & 13.0480 & 0.2126 \\
NL01   & SPADE     & 0.1361 & 0.3607 & 12.7894 & 0.2063 \\     
NL01   & None      & 0.1676 & 0.4022 & 13.6127 & 0.2332 \\
\hline
NL02   & Stack     & \textbf{0.1562} & \textbf{0.3880} & 13.0982 & \textbf{0.2210} \\
NL02   & AdaIN     & 0.1654 & 0.3998 & \textbf{13.0116} & 0.2311 \\
NL02   & Attention & 0.1658 & 0.4004 & 13.0685 & 0.2335 \\
NL02   & SPADE     & 0.1611 & 0.3943 & 13.2121 & 0.2263 \\     
NL02   & None      & 0.2630 & 0.5079 & 17.8102 & 0.3234 \\
\hline

NL03*   & Stack     & \textbf{0.1891} & \textbf{0.4264} & \textbf{33.4290} & \textbf{0.2709} \\
NL03*   & AdaIN     & 0.2040 & 0.4437 & 33.7215 & 0.2880 \\
NL03*   & Attention & 0.2113 & 0.4518 & 33.9402 & 0.2965 \\
NL03*   & SPADE     & 0.2182 & 0.4598 & 36.1949 & 0.3300 \\     
NL03*   & None      & 0.2687 & 0.5113 & 59.3721 & 0.4039 \\
\hline

NL03**   & Stack     & \textbf{0.1446} & \textbf{0.3714} & \textbf{35.3016} & \textbf{0.2140} \\
NL03**   & AdaIN     & 0.1475 & 0.3758 & 55.1334 & 0.2409 \\
NL03**   & Attention & 0.1668 & 0.4000 & 38.1286 & 0.2382 \\
NL03**   & SPADE     & 0.1687 & 0.4023 & 42.2689 & 0.2403 \\     
NL03**   & None      & 0.1619 & 0.3947 & 56.1288 & 0.2563 \\

\hline
\end{tabularx}
\caption{The quantitative results of different fusion methods.\normalfont{NL03* denotes NL03-0,1 and NL03** denotes NL03-2,3,4,5.}}
\label{sup:tab:fusion}
\end{table}

As shown in Table~\ref{sup:tab:fusion}, stack performs best in all situations especially with limited data (NL03). \sg{Cross} attention \sg{performs} relatively worse. 


\subsection{NL02: Switcher Module}
We use a switcher module to rotate the affine transformation, allowing the NL01 affine transformation matrices to adapt to NL02 characters. In this way, the NL01 pretrained parameters could be directly used to initiate the parameters of CAR when training on NL02 data, resulting in faster converage and better performance. The switcher module is actually a 6 × 6 diagonal matrix with all positions set to 0 except for (0,4), (4,0), (2,5), and (5,2), which have values of 1, as shown in Eq.~\ref{eq:switcher}. 

\begin{equation}
\label{eq:switcher}
    \begin{bmatrix}
  0 & 0 & 0 & 0 & 1 & 0 \\
  0 & 0 & 0 & 0 & 0 & 0 \\
  0 & 0 & 0 & 0 & 0 & 1 \\
  0 & 0 & 0 & 0 & 0 & 0 \\
  1 & 0 & 0 & 0 & 0 & 0 \\
  0 & 0 & 1 & 0 & 0 & 0
\end{bmatrix}
\end{equation}

\subsection{Iterative Invocation V.S. Three-Components CARs}
\begin{figure}[!ht]
    \centering
    \begin{subfigure}[b]{\linewidth}
        \centering
        \includegraphics[width=\linewidth]{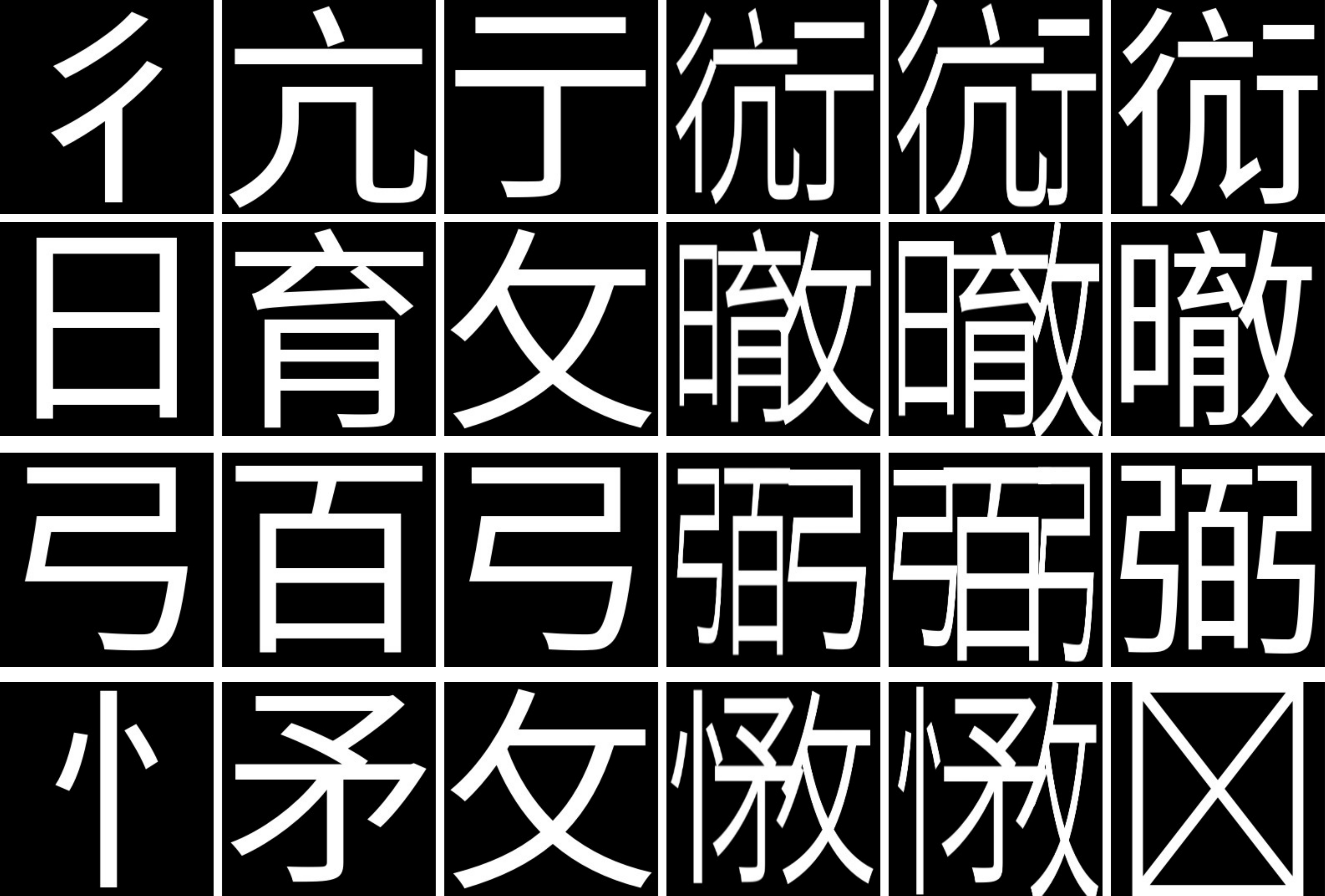}
        \caption{NL04}
        \label{fig:nl04}
    \end{subfigure}
    \begin{subfigure}[b]{\linewidth}
        \centering
        \includegraphics[width=\linewidth]{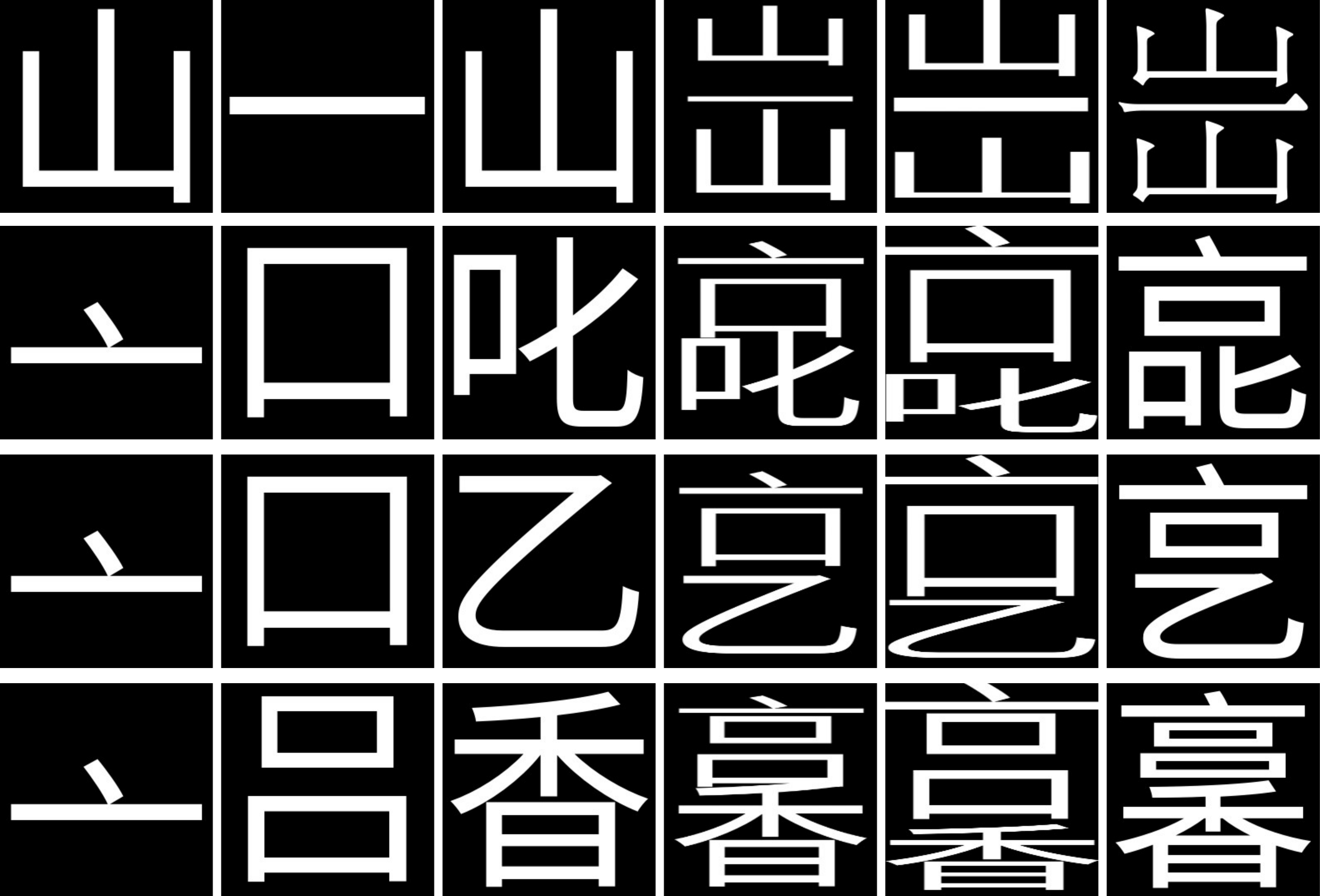}
        \caption{NL05}
        \label{fig:nl05}
    \end{subfigure}
    
    \caption{The NL04(\ref{fig:nl04}) NL05(\ref{fig:nl05}) samples are generated by two components CAR with iterative invocation and three components CAR. \normalfont{For each row, there are component 1, component 2, component 3, samples generated by two components CAR with iterative invocation strategy, samples generated by three components CAR, and ground truth, respectively. Ground truth displays ".notdef" character means the character is not supported by the font.}}
    \label{fig:iterative_vs_three_components}
\end{figure}

We \sg{present} some samples generated by two components CAR with the iterative invocation strategy and the three-components CAR to further illustrate the effectiveness of the approach, as shown in Figure~\ref{fig:iterative_vs_three_components}.
Compared with the three-components CAR, the iterative invocation strategy achieved better results. \jy{However, the performance of CAR on NL04 and NL05 still needs improvement.}


\subsection{Integration with Glyphs}
\begin{figure}[htb]
    \centering
    \includegraphics[width=\linewidth]{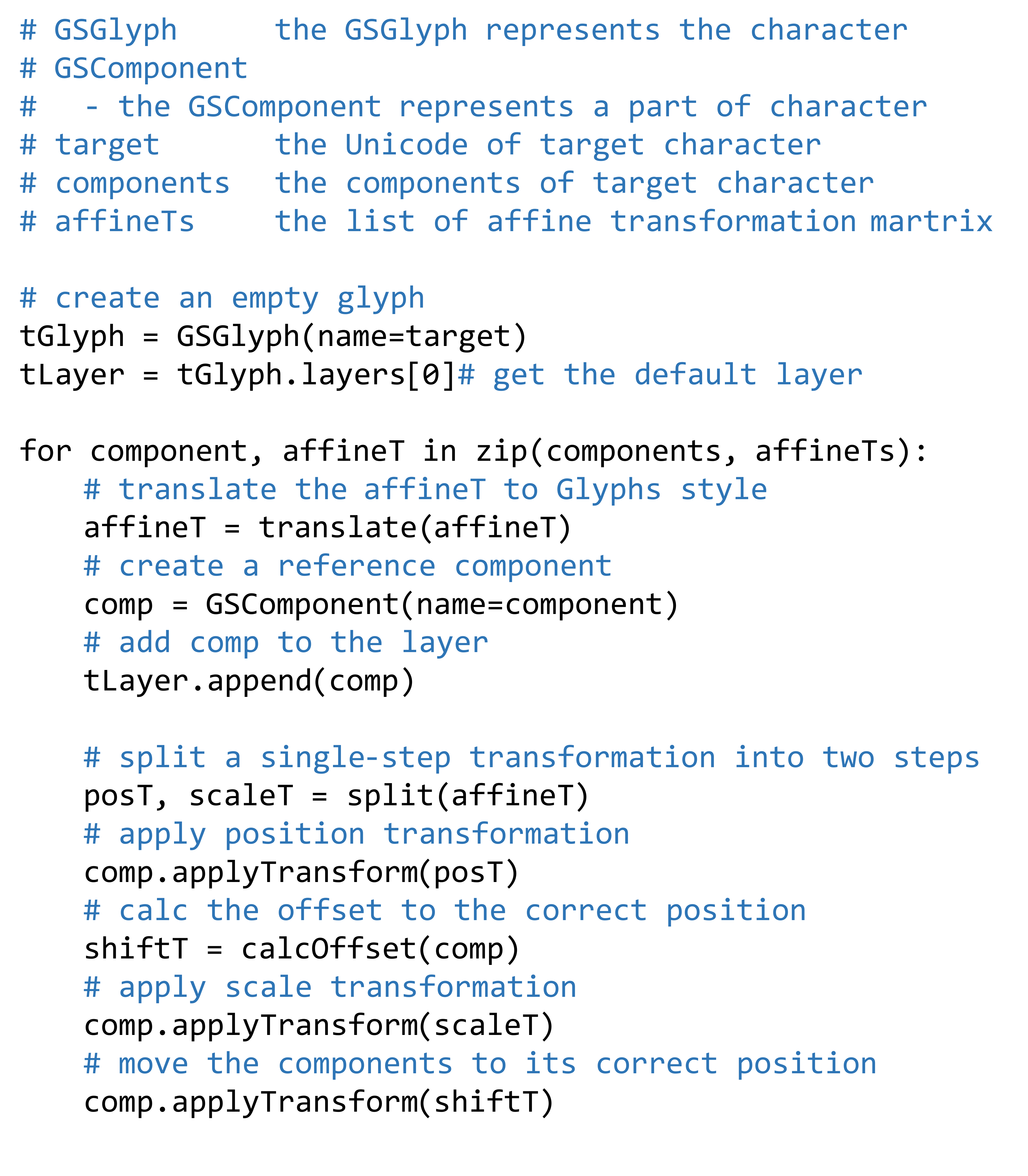}
    \caption{The python-like pseudocode for Glyphs script.}
    \label{fig:pseudocode_glyphs}
\end{figure}

As detailed in our paper, our method could generate high-quality Chinese vector fonts in large scale, which is expected to greatly improve the efficiency of Chinese font design. Therefore, we have developed a plug-in for Glyphs to enable designers to rapidly generate numerous style-consistent, editable characters based on carefully designed components. The pseudocode is shown in Fig.~\ref{fig:pseudocode_glyphs}.

For integrating our method into \textbf{Glyphs}, our script bridges the gap between PyTorch and Glyphs. There are three main differences between them. First, Glyphs and PyTorch's affine transformation parameters are different in order as shown in Eq.(\ref{eq:affine_order}). Note that Glyphs accepts a flattened affine transformation parameter. We manually change it to two rows here for comparison. Second, the effect of affine transformation parameters varies between PyTorch and Glyphs. PyTorch's scale parameters are the reciprocal of Glyphs', and the shift parameters in PyTorch are normalized to $[-2, 2]$, unlike in Glyphs. Third, the affine transformation origin is different: PyTorch uses image center while Glyphs uses the left-bottom corner of the component. 

\begin{equation}
    \begin{split}
    \label{eq:affine_order}
        A_{PyTorch} &= \begin{bmatrix} scale_x & skew_x & translation_x \\ skew_y & scale_y & translation_y \end{bmatrix} \\
        A_{Glyphs} &= \begin{bmatrix} scale_x & skew_x & skew_y \\ scale_y & translation_x & translation_y \end{bmatrix}
    \end{split}
\end{equation}
We additionally elaborate on the built-in classes in Glyphs Python Runtime\footnote{\url{https://docu.glyphsapp.com/}} that are used in our script. GSGlyph represents a character with at least one layer. The concept of the layer is the same as that in Photoshop. To create a character, components of the target characters, which are represented by GSComponent, are added to the layer sequentially. The GSComponent serves as a reference to the original character, synchronizing any modifications. Note that using GSPath (the Bezier curves) to migrate control points of components directly to the target character is also feasible. However, employing GSComponent enables designers to optimize components first and then release them into control points for rectifying, providing a more preferred workflow for designers.

%% file: supp-eval.tex
\section{Evaluation}
\subsection{Experimental Details}
\subsubsection{Preparing Data}
\begin{figure}[htb]
    \centering
    \includegraphics[width=0.4\linewidth]{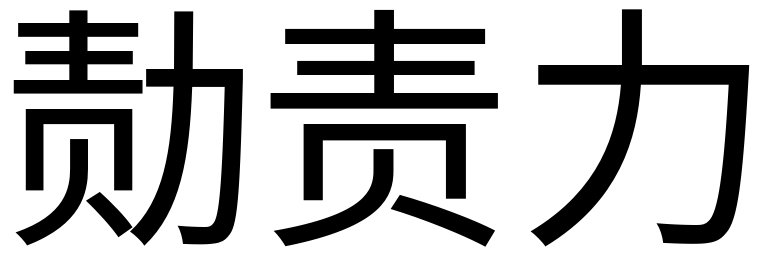}
    \caption{The examples of character-components images.}
    \label{fig:training_examples}
\end{figure}

Given a font, we extract all characters supported by the font, and we render each character and its components into $256\times 256$ images using PIL.ImageFont. We horizontally stack them (\ie the images of character and its components) together to form the character-components image, as shown in Fig.~\ref{fig:training_examples}. Note that we center the character and component in the images to ensure consistency because some components may be placed on the left or right by default. We classified these images by layout and used them as the dataset for specific layouts. In other words, our models are trained and evaluated on the character-components images that are extracted from one or multiple fonts and under the specific layout. We ensure repeatability of the data shuffle with the specified seed of 30.

\subsubsection{Hyperparameters}
\label{sec:hyperarameters}

\jy{We have performed grid search on the hyperparameters for each layout using SourceHanSans. Once determined, we used them in all other experiments for other fonts.} SGD momentum optimizer is used with its initial learning rate of $2e-3$ and momentum of 0.9. The StepLR scheduler was employed, reducing the learning rate by half every six epochs. We trained all models used in experiments for 42 epochs. The weights of different losses varied by layout. For stable layouts like NL01 and NL02, we only enabled the pixel and overlap loss. For NL03, enabling all losses is a must to guide the training in the correct direction. The weights we used in all experiments are listed in Tab.~\ref{tab:weight selection}. Note that we tested the selection of each loss weight for at least 5 times to obtain the final selections. 

\subsubsection{Miscellaneous}
We conducted all experiments on a NVIDIA 4090 with 24 GB of VRAM on Ubuntu 20.04. To obtain objective quantitative results, we ran each experiment three times and reported the average as the final result.
\begin{table}[ht]
\centering
\begin{tabular}{l|cccc}
\hline
Layout & Pixel & Overlap & Centroid & Inertia \\ \hline
NL01   & $1$           & $0$ or $1$         & $0$               & $0$              \\
NL02   & $1$            & $0$ or $1$         & $0$              & $0$              \\
NL03   & $1$            & $1$              & $5e-2$           & $1e-8$          \\ \hline

\end{tabular}
\caption{The weight of losses we used in our experiments.}
\label{tab:weight selection}
\end{table}



\subsection{Font Generation}
\subsubsection{Comparison with Component Composition GANs}
\begin{figure}[htb]
    \centering
    \includegraphics[width=0.95\linewidth]{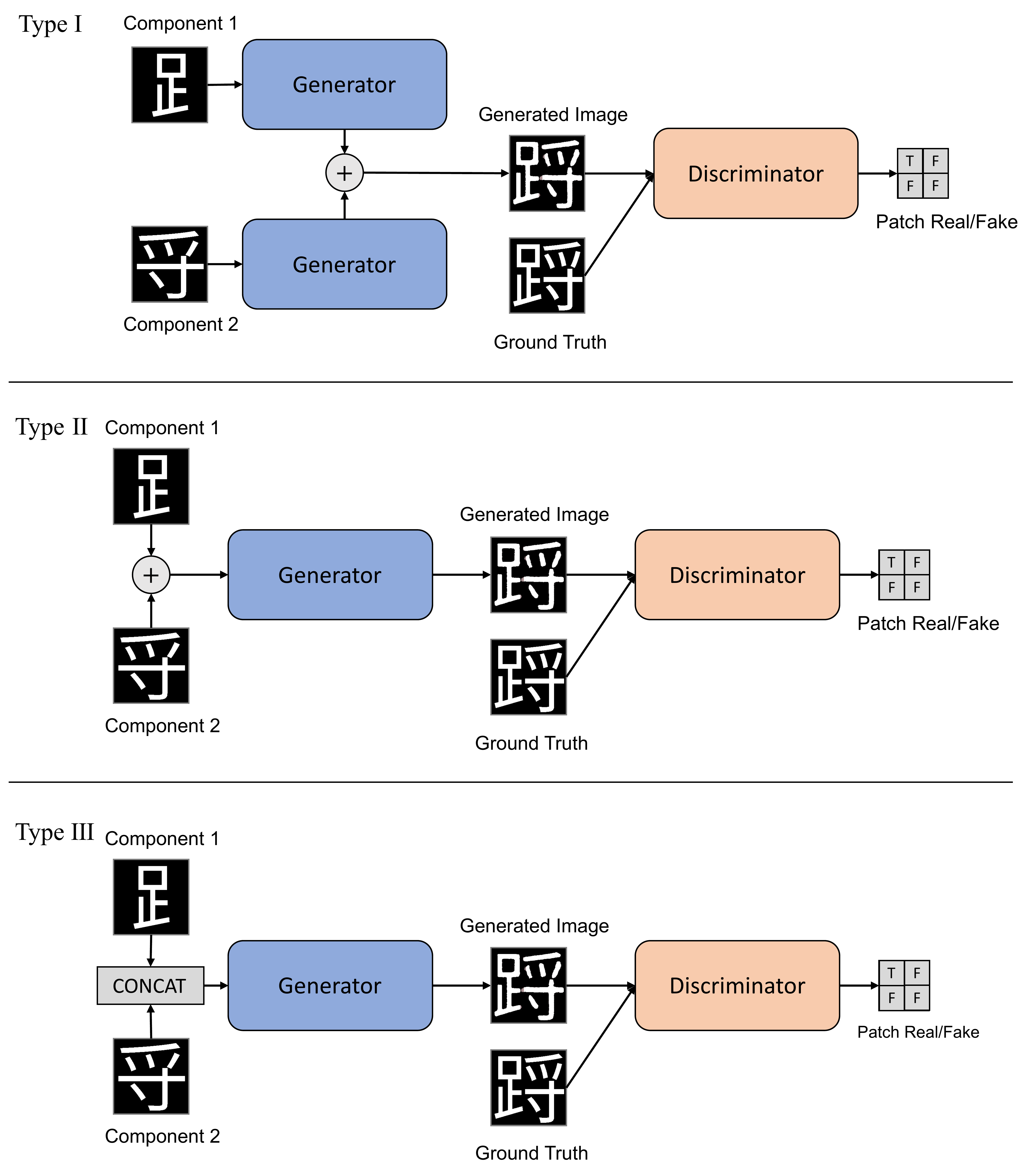}
    \caption{The architectures of GAN based on component composition.}
    \label{fig:gan_arch}
\end{figure}

In our main paper, we compared our CAR to three-component composition GANs. We trained our proposed CARs and self-implemented GANs on dataset extracted from SourceHanSans. The results are reported in the Tab.5 in our main paper. For clarity, we additionally detailed the architectures of GANs in Fig.~\ref{fig:gan_arch}. For the type-\uppercase\expandafter{\romannumeral1} , two separate generators process each component independently, with the transformed components summed to form the final output. In the second implementation, components are combined initially and input into a single generator for implicit transformation. For the type-\uppercase\expandafter{\romannumeral3},  component images are concatenated in the channel dimension, with the remaining process similar to Type-\uppercase\expandafter{\romannumeral2}. The generator and discriminator implementations are adapted from a Pix2Pix~\footnote{\url{https://github.com/eriklindernoren/PyTorch-GAN/blob/master/implementations/pix2pix/models.py}} implementation. Source code and additional details are provided in the supplementary material.

\subsubsection{Comparison with the SOTA}

\begin{figure}[ht]
    \centering
    \includegraphics[width=\linewidth]{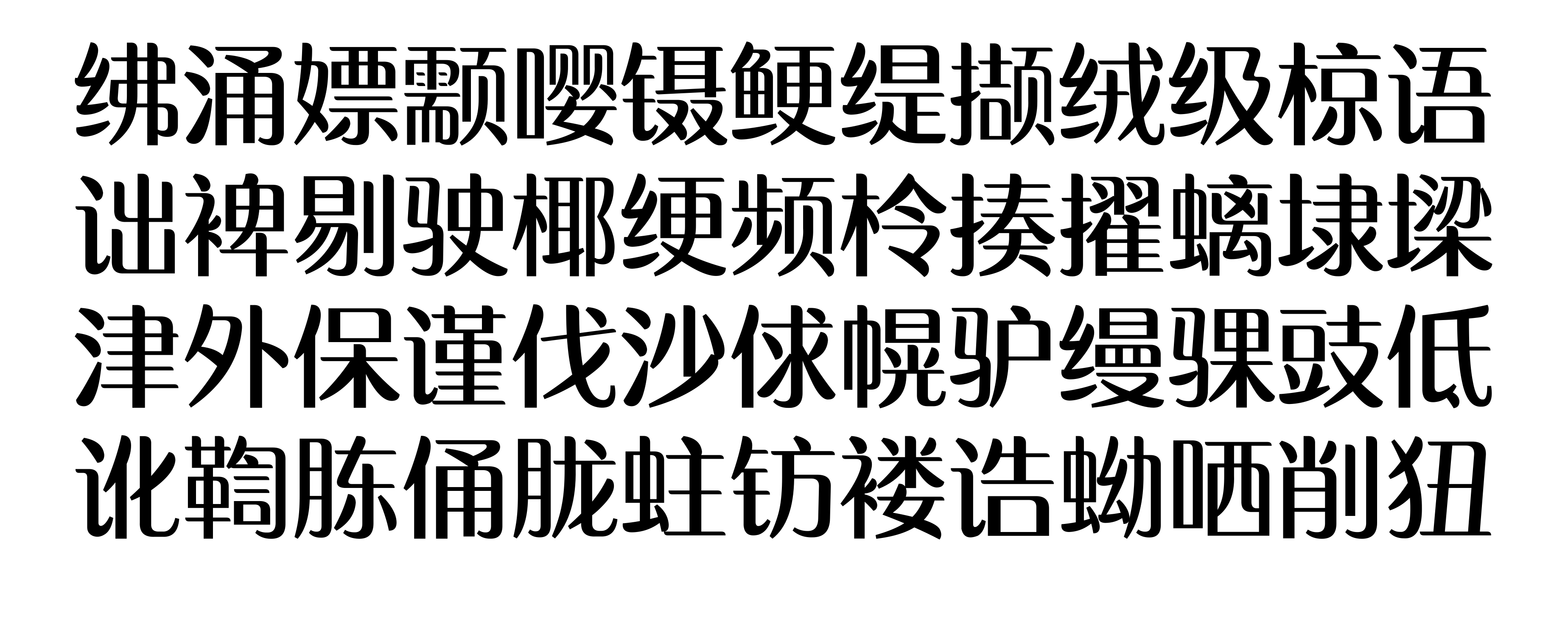}
    \caption{The 52 characters rendered with FZSYJ.}
    \label{fig:dvf_v2_52chars}
\end{figure}

\begin{figure}[ht]
    \centering
    \includegraphics[width=\linewidth]{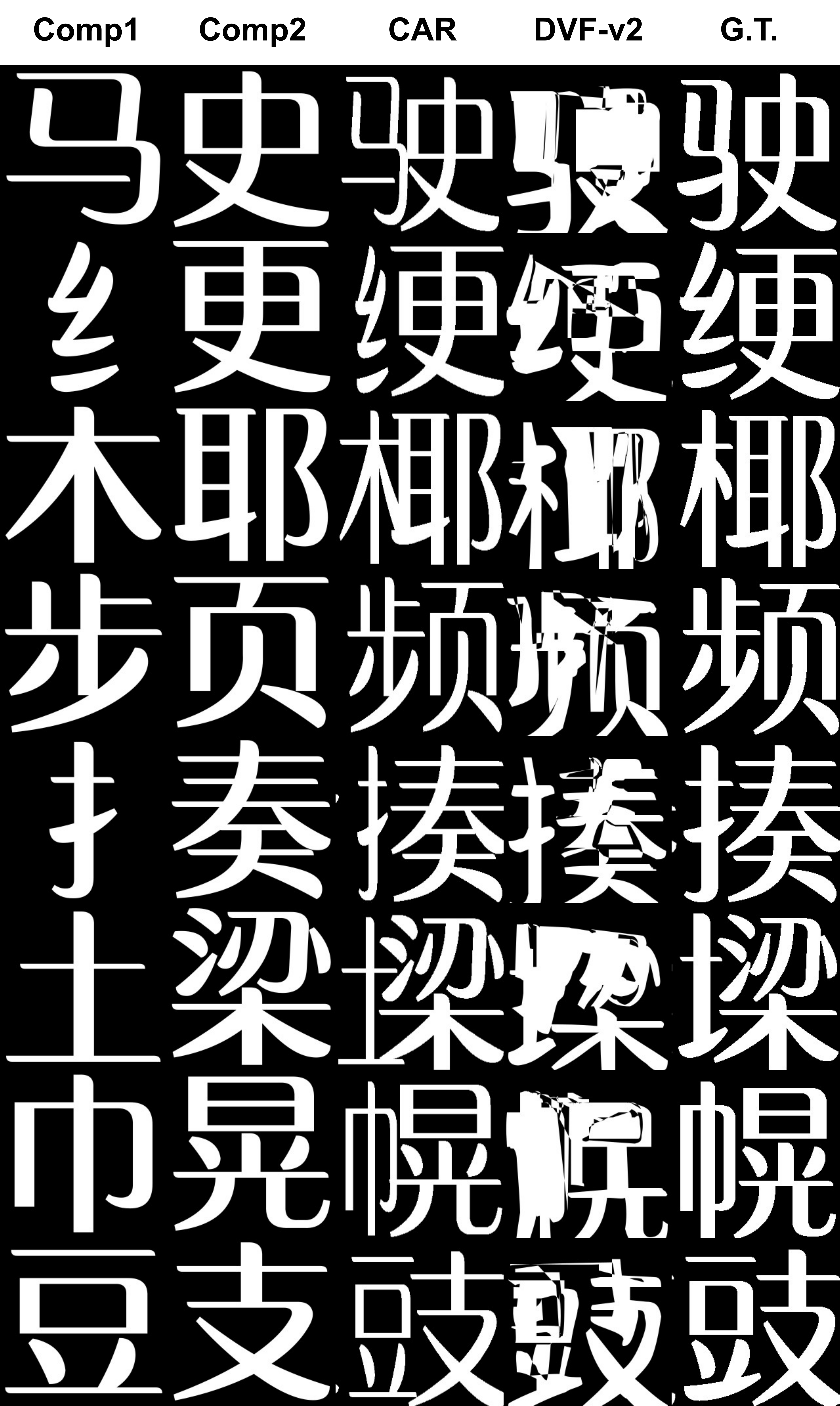}
    \caption{More results produced by CAR and DVF-v2.}
    \label{fig:more_results_dvf_car}
\end{figure}

In our paper, we compared our method with the SOTA vector font generation method, DeepVecFont-v2~\cite{wang2023deepvecfont} (abbr. DVF-v2). DVF-v2 generates vector font based on style transfer, while ours generates vector font via component composition. Nevertheless, since both methods tackle the task of vector font generation, we believe it is necessary to compare the method with the SOTA to evaluate our performance. For DVF-v2, we followed the training procedure provided in their source code. We trained DVF-v2 on 210 fonts and tested it on the target font, FangZhengShuiYunJian (abbr. FZSYJ). All fonts used are provided by DVF-v2 in their repository. Note that in DVF-v2, it is necessary to define which characters the model needs to generate in advance. Thus, we randomly selected 52 characters, as shown in Fig~\ref{fig:dvf_v2_52chars}, that are included in the \textit{test set} when training CARs. The DVF-v2 used for comparison was trained for 750 epochs. For CAR, we trained our CAR on the NL01 training set of FZSJY, which includes 2833 characters, and evaluated it on those 52 characters mentioned above. The hyperparameters and training details follow the description in Sec.~\ref{sec:hyperarameters}. The comparison of hyperparamers between DVF-v2 and ours is listed in Tab.~\ref{tab:hyperparam_dvf_car}. The quantitative results and generated samples are reported in the main paper. We additionally show more generated results produced by both methods in Fig.~\ref{fig:more_results_dvf_car}.


\begin{table}[ht]
\small
\begin{tabularx}{\linewidth}{l|XX}
\hline
Hyperparams & DVF-v2               & CAR (Ours)         \\ \hline
train on       & 210 fonts $\times$ 52 chars & FZSYJ (2833 chars) \\
test on        & FZSYJ (52 chars)     & FZSYJ (52 chars)   \\
epoch          & 750                  & 42                 \\
ref\_nshot     & 8                    & N/A                \\
max\_seq\_len  & 231                  & No Limitation      \\
lr             & $1e-4$               & $2e-3$               \\ \hline  
\end{tabularx}
\caption{The comparison of hyperparameters between DVF-v2 and CAR.}
\label{tab:hyperparam_dvf_car}
\end{table}

\subsection{Zero Shot Font Extension}
\begin{figure}[ht]
    \centering
    \includegraphics[width=\linewidth]{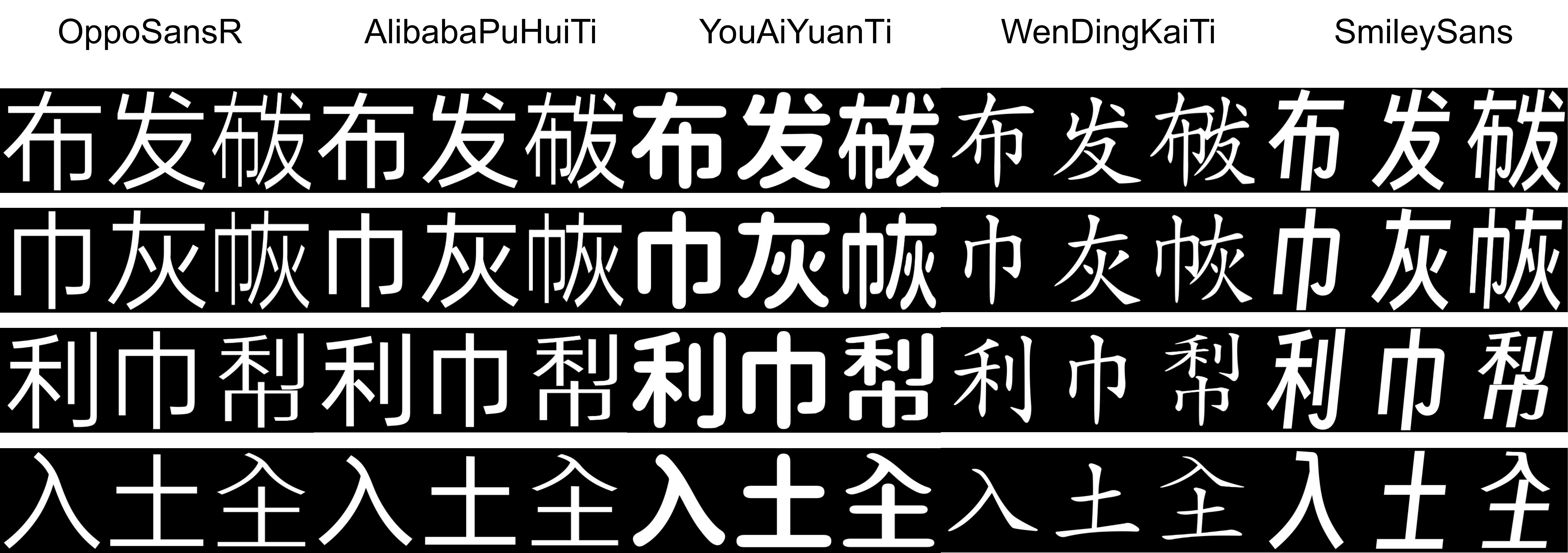}
    \caption{The zero-shot generated results of OppoSansR, AlibabaPuHuiTi, YouAiYuanTi, WenDingKaiTi, and SmileySans. The model was trained on SourceHanSans. The generated characters are all in CJK extension F. }
    \label{fig:zero-shot-generated-samples}
\end{figure}

Our method could serve as an out-of-the-box font extension tool for multiple fonts with varying styles in a zero-shot manner. That means the CAR trained on one font could synthesize the characters that other fonts did not support before, without being finetuned on those fonts, to extend their coverage of Chinese characters. More zero-shot font generation results are shown in Fig.~\ref{fig:zero-shot-generated-samples}. 

\subsection{More Results}
\begin{figure}[H]
    \centering
    \includegraphics[width=\linewidth]{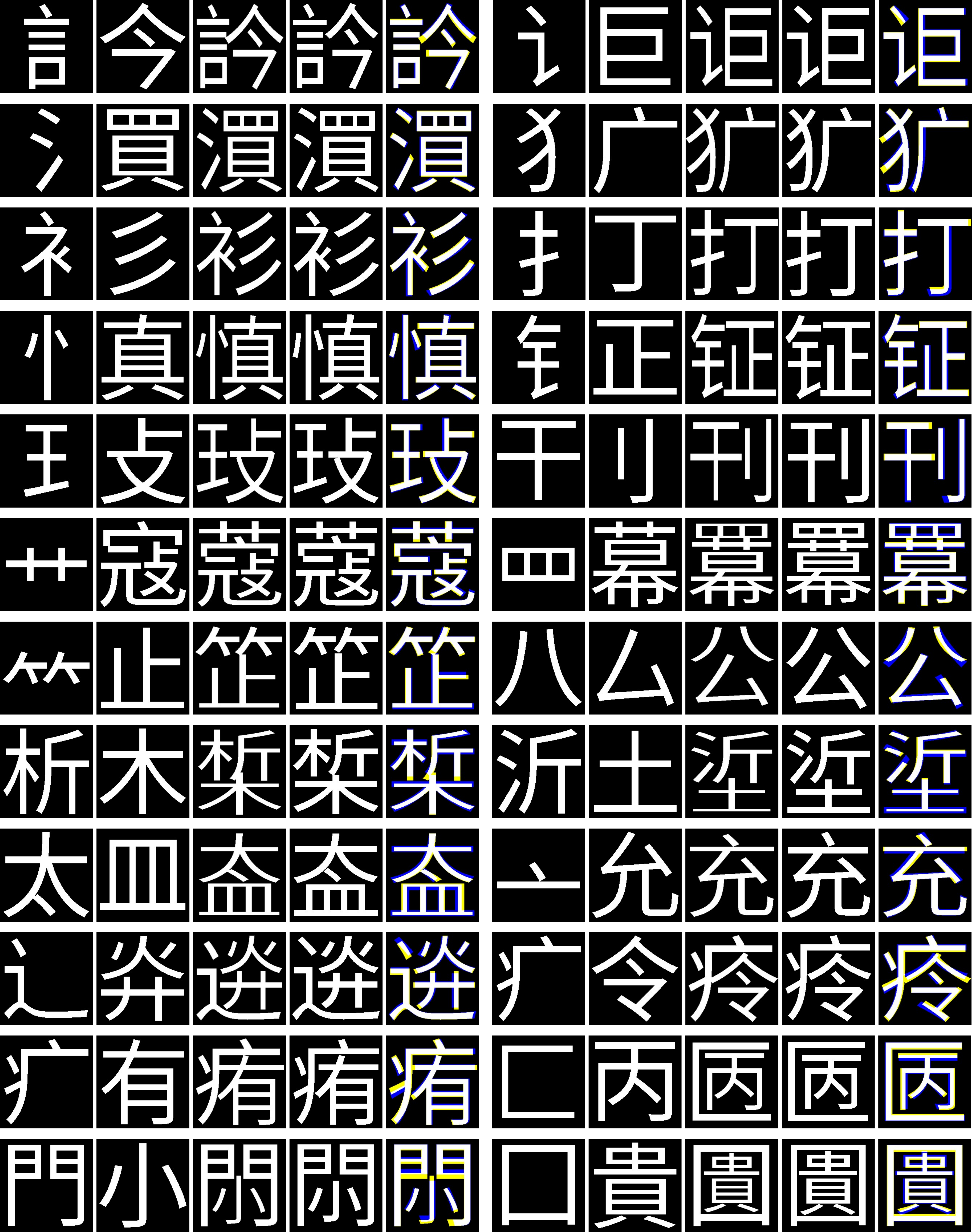}
    \caption{Samples generated by our method. The figure displays component 1, component 2, generated sample, ground truth, and the difference between generated sample and ground truth, respectively.}
    \label{fig:moreresults}
\end{figure}

We show more generated results of CAR for NL01, NL02, and NL03 in Fig.~\ref{fig:moreresults}. The generation quality demonstrates the effectiveness of our approach.

\subsection{User Study}
We provide further details on two user studies conducted during the evaluation.

\subsubsection{Font Generation}
\begin{figure}[H]
    \centering
    \includegraphics[width=0.98\linewidth]{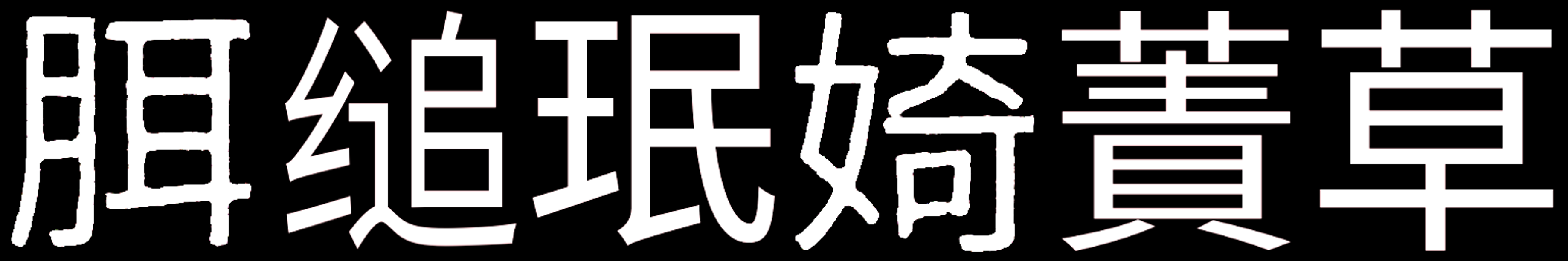}
    \caption{Images for font generation user study. \normalfont{From left to right, the 1st and 4th images were generated by GANs, the 2nd and 5th images were generated by our method, and the 3rd and 6th images were designed by professional designers.}}
    \label{fig:user-study-font-generation}
\end{figure}

We asked participants six questions, each inquiring if they believed the displayed image was algorithm-generated. Of the 6 images we prepare, 2 are GAN-generated character images, 2 are images generated by our method, and 2 are real character images. Figure~\ref{fig:user-study-font-generation} presents the images used, with No.1 and No.4 generated by GANs, No.2 and No.5 by our method, and No.3 and No.6 by professional designers. To minimize interference, participants viewed one question at a time. 

\subsubsection{Font Extension} 
\begin{figure}[H]
    \centering
    \includegraphics[width=\linewidth]{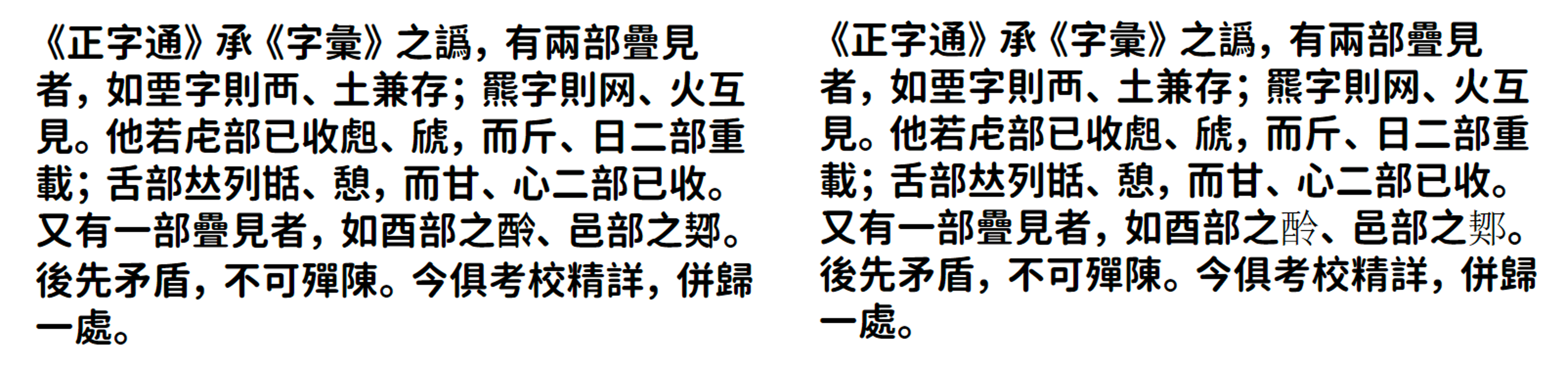}
    \caption{Images for font completion user study. \textbf{Left}: Rendered by the completed font. \textbf{Right}: Rendered by the original font.}
    \label{fig:user-study-font-completion}
\end{figure}

Participants were asked to select the item exhibiting greater visual consistency from two prepared options. The images provided to users are displayed in Figure~\ref{fig:user-study-font-completion}. Our method demonstrated a clear advantage, validating its effectiveness. It is important to note that the extended font was generated in a zero-shot manner. In other words, our model for font extended was trained only on SourceHanSans and not fine-tuned on the specific font, further highlighting its generalization capabilities.

%% file: supp-impact.tex
\section{Paradigm Shift of Chinese Font Design}

Our method has the potential to establish new paradigms in Chinese font design. As shown in Fig.~\ref{fig:paradigm_shift}, in traditional font design processes, manual character composition (the red block) consumes a significant portion of time. This is also the reason why current Chinese fonts typically cover only 6,763 characters and struggle to encompass the entire Chinese character set. However, this step can be replaced by the automatic character composition (the purple blocks) introduced by our method. Transitioning from manual to automated, our approach not only achieves a significant breakthrough in both efficiency and scalability but also makes the long-standing goal of "enabling widespread support for a more comprehensive Chinese character set" in typefaces a possibility.



%% file: ijcai24.bbl
\begin{thebibliography}{}

\bibitem[\protect\citeauthoryear{Aoki and Aizawa}{2022}]{aoki2022svg}
Haruka Aoki and Kiyoharu Aizawa.
\newblock Svg vector font generation for chinese characters with transformer.
\newblock In {\em 2022 IEEE International Conference on Image Processing (ICIP)}, pages 646--650. IEEE, 2022.

\bibitem[\protect\citeauthoryear{Campbell and Kautz}{2014}]{campbell2014learning}
Neill~DF Campbell and Jan Kautz.
\newblock Learning a manifold of fonts.
\newblock {\em ACM Transactions on Graphics (ToG)}, 33(4):1--11, 2014.

\bibitem[\protect\citeauthoryear{Carlier \bgroup \em et al.\egroup }{2020}]{carlier2020deepsvg}
Alexandre Carlier, Martin Danelljan, Alexandre Alahi, and Radu Timofte.
\newblock Deepsvg: A hierarchical generative network for vector graphics animation.
\newblock {\em Advances in Neural Information Processing Systems}, 33:16351--16361, 2020.

\bibitem[\protect\citeauthoryear{Consortium}{2016}]{unicode9.0}
Unicode Consortium.
\newblock {\em The Unicode Standard}.
\newblock Unicode Consortium, Mountain View, CA, 9.0 edition, July 2016.
\newblock Pages 689--692.

\bibitem[\protect\citeauthoryear{Fan \bgroup \em et al.\egroup }{2017}]{fan2017point}
Haoqiang Fan, Hao Su, and Leonidas~J Guibas.
\newblock A point set generation network for 3d object reconstruction from a single image.
\newblock In {\em Proceedings of the IEEE conference on computer vision and pattern recognition}, pages 605--613, 2017.

\bibitem[\protect\citeauthoryear{Gao \bgroup \em et al.\egroup }{2019a}]{gao2019automatic}
Yichen Gao, Zhouhui Lian, Yingmin Tang, and Jianguo Xiao.
\newblock Automatic generation of chinese vector fonts via deep layout inferring.
\newblock In {\em SIGGRAPH Asia 2019 Technical Briefs}, pages 33--36, 2019.

\bibitem[\protect\citeauthoryear{Gao \bgroup \em et al.\egroup }{2019b}]{gao2019artistic}
Yue Gao, Yuan Guo, Zhouhui Lian, Yingmin Tang, and Jianguo Xiao.
\newblock Artistic glyph image synthesis via one-stage few-shot learning.
\newblock {\em ACM Transactions on Graphics (TOG)}, 38(6):1--12, 2019.

\bibitem[\protect\citeauthoryear{Goodfellow \bgroup \em et al.\egroup }{2014}]{goodfellowNIPS2014generative}
Ian Goodfellow, Jean Pouget-Abadie, Mehdi Mirza, Bing Xu, David Warde-Farley, Sherjil Ozair, Aaron Courville, and Yoshua Bengio.
\newblock Generative adversarial nets.
\newblock In Z.~Ghahramani, M.~Welling, C.~Cortes, N.~Lawrence, and K.Q. Weinberger, editors, {\em Advances in Neural Information Processing Systems}, volume~27. Curran Associates, Inc., 2014.

\bibitem[\protect\citeauthoryear{Guo and Kunii}{1991}]{guo1991modeling}
Qinglian Guo and Tosiyasu~L Kunii.
\newblock Modeling the diffuse painting of ‘sumie’.
\newblock {\em Modeling in Computer Graphics}, pages 329--338, 1991.

\bibitem[\protect\citeauthoryear{Guo \bgroup \em et al.\egroup }{2018}]{guo2018creating}
Yuan Guo, Zhouhui Lian, Yingmin Tang, and Jianguo Xiao.
\newblock Creating new chinese fonts based on manifold learning and adversarial networks.
\newblock In {\em Eurographics (Short Papers)}, pages 61--64, 2018.

\bibitem[\protect\citeauthoryear{He \bgroup \em et al.\egroup }{2022}]{he2022diff}
Haibin He, Xinyuan Chen, Chaoyue Wang, Juhua Liu, Bo~Du, Dacheng Tao, and Yu~Qiao.
\newblock Diff-font: Diffusion model for robust one-shot font generation.
\newblock {\em arXiv preprint arXiv:2212.05895}, 2022.

\bibitem[\protect\citeauthoryear{Heusel \bgroup \em et al.\egroup }{2017}]{heusel2017gans}
Martin Heusel, Hubert Ramsauer, Thomas Unterthiner, Bernhard Nessler, and Sepp Hochreiter.
\newblock Gans trained by a two time-scale update rule converge to a local nash equilibrium.
\newblock {\em Advances in neural information processing systems}, 30, 2017.

\bibitem[\protect\citeauthoryear{Ho \bgroup \em et al.\egroup }{2020}]{ho2020denoising}
Jonathan Ho, Ajay Jain, and Pieter Abbeel.
\newblock Denoising diffusion probabilistic models.
\newblock {\em Advances in Neural Information Processing Systems}, 33:6840--6851, 2020.

\bibitem[\protect\citeauthoryear{Isola \bgroup \em et al.\egroup }{2017}]{isola2017image}
Phillip Isola, Jun-Yan Zhu, Tinghui Zhou, and Alexei~A Efros.
\newblock Image-to-image translation with conditional adversarial networks.
\newblock In {\em Proceedings of the IEEE conference on computer vision and pattern recognition}, pages 1125--1134, 2017.

\bibitem[\protect\citeauthoryear{Jaderberg \bgroup \em et al.\egroup }{2015}]{STN_NIPS2015_33ceb07b}
Max Jaderberg, Karen Simonyan, Andrew Zisserman, and koray kavukcuoglu.
\newblock Spatial transformer networks.
\newblock In C.~Cortes, N.~Lawrence, D.~Lee, M.~Sugiyama, and R.~Garnett, editors, {\em Advances in Neural Information Processing Systems}, volume~28. Curran Associates, Inc., 2015.

\bibitem[\protect\citeauthoryear{Jiang \bgroup \em et al.\egroup }{2019}]{jiang2019scfont}
Yue Jiang, Zhouhui Lian, Yingmin Tang, and Jianguo Xiao.
\newblock Scfont: Structure-guided chinese font generation via deep stacked networks.
\newblock In {\em Proceedings of the AAAI conference on artificial intelligence}, volume~33, pages 4015--4022, 2019.

\bibitem[\protect\citeauthoryear{Kingma and Welling}{2013}]{kingma2013auto}
Diederik~P Kingma and Max Welling.
\newblock Auto-encoding variational bayes.
\newblock {\em arXiv preprint arXiv:1312.6114}, 2013.

\bibitem[\protect\citeauthoryear{Kong \bgroup \em et al.\egroup }{2022}]{kong2022look}
Yuxin Kong, Canjie Luo, Weihong Ma, Qiyuan Zhu, Shenggao Zhu, Nicholas Yuan, and Lianwen Jin.
\newblock Look closer to supervise better: One-shot font generation via component-based discriminator.
\newblock In {\em Proceedings of the IEEE/CVF Conference on Computer Vision and Pattern Recognition}, pages 13482--13491, 2022.

\bibitem[\protect\citeauthoryear{Lee}{1999}]{lee1999simulating}
Jintae Lee.
\newblock Simulating oriental black-ink painting.
\newblock {\em IEEE Computer Graphics and Applications}, 19(3):74--81, 1999.

\bibitem[\protect\citeauthoryear{Lian and Gao}{2022}]{lian2022cvfont}
Zhouhui Lian and Yichen Gao.
\newblock Cvfont: Synthesizing chinese vector fonts via deep layout inferring.
\newblock In {\em Computer Graphics Forum}, volume~41, pages 212--225. Wiley Online Library, 2022.

\bibitem[\protect\citeauthoryear{Liu and Lian}{2023}]{liu2023fonttransformer}
Yitian Liu and Zhouhui Lian.
\newblock Fonttransformer: Few-shot high-resolution chinese glyph image synthesis via stacked transformers.
\newblock {\em Pattern Recognition}, page 109593, 2023.

\bibitem[\protect\citeauthoryear{Lopes \bgroup \em et al.\egroup }{2019}]{lopes2019learned}
Raphael~Gontijo Lopes, David Ha, Douglas Eck, and Jonathon Shlens.
\newblock A learned representation for scalable vector graphics.
\newblock In {\em Proceedings of the IEEE/CVF International Conference on Computer Vision}, pages 7930--7939, 2019.

\bibitem[\protect\citeauthoryear{Lyu \bgroup \em et al.\egroup }{2017}]{lyu2017auto}
Pengyuan Lyu, Xiang Bai, Cong Yao, Zhen Zhu, Tengteng Huang, and Wenyu Liu.
\newblock Auto-encoder guided gan for chinese calligraphy synthesis.
\newblock In {\em 2017 14th IAPR International Conference on Document Analysis and Recognition (ICDAR)}, volume~1, pages 1095--1100. IEEE, 2017.

\bibitem[\protect\citeauthoryear{Park \bgroup \em et al.\egroup }{2019}]{park2019semantic}
Taesung Park, Ming-Yu Liu, Ting-Chun Wang, and Jun-Yan Zhu.
\newblock Semantic image synthesis with spatially-adaptive normalization.
\newblock In {\em Proceedings of the IEEE/CVF conference on computer vision and pattern recognition}, pages 2337--2346, 2019.

\bibitem[\protect\citeauthoryear{Park \bgroup \em et al.\egroup }{2021}]{park2021multiple}
Song Park, Sanghyuk Chun, Junbum Cha, Bado Lee, and Hyunjung Shim.
\newblock Multiple heads are better than one: Few-shot font generation with multiple localized experts.
\newblock In {\em Proceedings of the IEEE/CVF International Conference on Computer Vision}, pages 13900--13909, 2021.

\bibitem[\protect\citeauthoryear{Reddy \bgroup \em et al.\egroup }{2021}]{reddy2021im2vec}
Pradyumna Reddy, Michael Gharbi, Michal Lukac, and Niloy~J Mitra.
\newblock Im2vec: Synthesizing vector graphics without vector supervision.
\newblock In {\em Proceedings of the IEEE/CVF Conference on Computer Vision and Pattern Recognition}, pages 7342--7351, 2021.

\bibitem[\protect\citeauthoryear{Sohl-Dickstein \bgroup \em et al.\egroup }{2015}]{pmlr-v37-sohl-dickstein15}
Jascha Sohl-Dickstein, Eric Weiss, Niru Maheswaranathan, and Surya Ganguli.
\newblock Deep unsupervised learning using nonequilibrium thermodynamics.
\newblock In Francis Bach and David Blei, editors, {\em Proceedings of the 32nd International Conference on Machine Learning}, volume~37 of {\em Proceedings of Machine Learning Research}, pages 2256--2265, Lille, France, 07--09 Jul 2015. PMLR.

\bibitem[\protect\citeauthoryear{Strassmann}{1986}]{strassmann1986hairy}
Steve Strassmann.
\newblock Hairy brushes.
\newblock {\em ACM Siggraph Computer Graphics}, 20(4):225--232, 1986.

\bibitem[\protect\citeauthoryear{Sun \bgroup \em et al.\egroup }{2018}]{sun2018pyramid}
Donghui Sun, Qing Zhang, and Jun Yang.
\newblock Pyramid embedded generative adversarial network for automated font generation.
\newblock In {\em 2018 24th International Conference on Pattern Recognition (ICPR)}, pages 976--981. IEEE, 2018.

\bibitem[\protect\citeauthoryear{Suveeranont and Igarashi}{2010}]{suveeranont2010example}
Rapee Suveeranont and Takeo Igarashi.
\newblock Example-based automatic font generation.
\newblock In {\em Smart Graphics: 10th International Symposium on Smart Graphics, Banff, Canada, June 24-26, 2010 Proceedings 10}, pages 127--138. Springer, 2010.

\bibitem[\protect\citeauthoryear{Tang \bgroup \em et al.\egroup }{2019}]{tang2019fontrnn}
Shusen Tang, Zeqing Xia, Zhouhui Lian, Yingmin Tang, and Jianguo Xiao.
\newblock Fontrnn: Generating large-scale chinese fonts via recurrent neural network.
\newblock In {\em Computer Graphics Forum}, volume~38, pages 567--577. Wiley Online Library, 2019.

\bibitem[\protect\citeauthoryear{Tian}{2017}]{tian2017zi2zi}
Yuchen Tian.
\newblock zi2zi: Master chinese calligraphy with conditional adversarial networks.
\newblock {\em Internet] https://github. com/kaonashi-tyc/zi2zi}, 3:2, 2017.

\bibitem[\protect\citeauthoryear{Vaswani \bgroup \em et al.\egroup }{2017}]{vaswani2017attention}
Ashish Vaswani, Noam Shazeer, Niki Parmar, Jakob Uszkoreit, Llion Jones, Aidan~N Gomez, {\L}ukasz Kaiser, and Illia Polosukhin.
\newblock Attention is all you need.
\newblock {\em Advances in neural information processing systems}, 30, 2017.

\bibitem[\protect\citeauthoryear{Wang and Lian}{2021}]{wang2021deepvecfont}
Yizhi Wang and Zhouhui Lian.
\newblock Deepvecfont: Synthesizing high-quality vector fonts via dual-modality learning.
\newblock {\em ACM Transactions on Graphics (TOG)}, 40(6):1--15, 2021.

\bibitem[\protect\citeauthoryear{Wang \bgroup \em et al.\egroup }{2020}]{wang2020attribute2font}
Yizhi Wang, Yue Gao, and Zhouhui Lian.
\newblock Attribute2font: Creating fonts you want from attributes.
\newblock {\em ACM Transactions on Graphics (TOG)}, 39(4):69--1, 2020.

\bibitem[\protect\citeauthoryear{Wang \bgroup \em et al.\egroup }{2022}]{wang2022aesthetic}
Yizhi Wang, Guo Pu, Wenhan Luo, Yexin Wang, Pengfei Xiong, Hongwen Kang, and Zhouhui Lian.
\newblock Aesthetic text logo synthesis via content-aware layout inferring.
\newblock In {\em Proceedings of the IEEE/CVF Conference on Computer Vision and Pattern Recognition}, pages 2436--2445, 2022.

\bibitem[\protect\citeauthoryear{Wang \bgroup \em et al.\egroup }{2023}]{wang2023deepvecfont}
Yuqing Wang, Yizhi Wang, Longhui Yu, Yuesheng Zhu, and Zhouhui Lian.
\newblock Deepvecfont-v2: Exploiting transformers to synthesize vector fonts with higher quality.
\newblock {\em arXiv preprint arXiv:2303.14585}, 2023.

\bibitem[\protect\citeauthoryear{Wong and Hsu}{1995}]{wong1995designing}
Paul Yiu~Chung Wong and Siu~Chi Hsu.
\newblock Designing chinese typeface using components.
\newblock In {\em Proceedings Nineteenth Annual International Computer Software and Applications Conference (COMPSAC'95)}, pages 416--421. IEEE, 1995.

\bibitem[\protect\citeauthoryear{Xia \bgroup \em et al.\egroup }{2023}]{xia2023vecfontsdf}
Zeqing Xia, Bojun Xiong, and Zhouhui Lian.
\newblock Vecfontsdf: Learning to reconstruct and synthesize high-quality vector fonts via signed distance functions.
\newblock {\em arXiv preprint arXiv:2303.12675}, 2023.

\bibitem[\protect\citeauthoryear{Xu \bgroup \em et al.\egroup }{2005}]{xu2005automatic}
Songhua Xu, Francis~CM Lau, William~K Cheung, and Yunhe Pan.
\newblock Automatic generation of artistic chinese calligraphy.
\newblock {\em IEEE Intelligent Systems}, 20(3):32--39, 2005.

\bibitem[\protect\citeauthoryear{Zhang \bgroup \em et al.\egroup }{2017}]{zhang2017drawing}
Xu-Yao Zhang, Fei Yin, Yan-Ming Zhang, Cheng-Lin Liu, and Yoshua Bengio.
\newblock Drawing and recognizing chinese characters with recurrent neural network.
\newblock {\em IEEE transactions on pattern analysis and machine intelligence}, 40(4):849--862, 2017.

\bibitem[\protect\citeauthoryear{Zhang \bgroup \em et al.\egroup }{2018a}]{zhang2018unreasonable}
Richard Zhang, Phillip Isola, Alexei~A Efros, Eli Shechtman, and Oliver Wang.
\newblock The unreasonable effectiveness of deep features as a perceptual metric.
\newblock In {\em Proceedings of the IEEE conference on computer vision and pattern recognition}, pages 586--595, 2018.

\bibitem[\protect\citeauthoryear{Zhang \bgroup \em et al.\egroup }{2018b}]{zhang2018separating}
Yexun Zhang, Ya~Zhang, and Wenbin Cai.
\newblock Separating style and content for generalized style transfer.
\newblock In {\em Proceedings of the IEEE conference on computer vision and pattern recognition}, pages 8447--8455, 2018.

\bibitem[\protect\citeauthoryear{Zhang \bgroup \em et al.\egroup }{2020}]{zhang2020ssnet}
Jianwei Zhang, Danni Chen, Guoqiang Han, Guanzhao Li, Junting He, Zhenmei Liu, and Zhihui Ruan.
\newblock Ssnet: Structure-semantic net for chinese typography generation based on image translation.
\newblock {\em Neurocomputing}, 371:15--26, 2020.

\bibitem[\protect\citeauthoryear{Zhou \bgroup \em et al.\egroup }{2011}]{zhou2011easy}
Baoyao Zhou, Weihong Wang, and Zhanghui Chen.
\newblock Easy generation of personal chinese handwritten fonts.
\newblock In {\em 2011 IEEE international conference on multimedia and expo}, pages 1--6. IEEE, 2011.

\end{thebibliography}
